\pdfoutput=1

\RequirePackage{fix-cm}
\documentclass{article}

\PassOptionsToPackage{table}{xcolor}

\usepackage{preprint,times}
\usepackage[T1]{fontenc}
\usepackage{microtype}

\usepackage{xcolor}
\usepackage{amssymb}
\usepackage{mathtools}

\usepackage{amsmath,amsfonts,bm}

\usepackage{algorithm}
\usepackage{algpseudocode}

\usepackage{graphicx}
\usepackage{array}
\usepackage{tabularx}
\usepackage{multirow}
\usepackage{makecell}
\usepackage{booktabs}
\usepackage{arydshln}
\usepackage{caption}
\usepackage{placeins}

\usepackage{pifont}
\usepackage{titletoc}

\usepackage{listings}
\usepackage{tcolorbox}
\tcbuselibrary{listings,breakable,skins}

\usepackage{url}
\usepackage{hyperref}

\hypersetup{
    colorlinks=true,
    linkcolor=red,
    citecolor=cyan,
    urlcolor=blue,
    pdftitle={Semantically Aligned KV Cache Compression for Efficient Agentic Autoregressive Video Generation},
    pdfauthor={Weizhong Huang, Huaijia Kou, Tianyu Xie, Xiawu Zheng, Liujuan Cao}
}

\definecolor{softcyan}{HTML}{CDEFF1}
\definecolor{softgray}{HTML}{F2F2F2}

\definecolor{promptTealBlue}{cmyk}{0.86,0,0.34,0.02}
\definecolor{promptPeriwinkle}{cmyk}{0.57,0.55,0,0}
\definecolor{promptForestGreen}{cmyk}{0.91,0,0.88,0.12}
\definecolor{promptMelon}{cmyk}{0,0.46,0.50,0}
\definecolor{promptshadow}{RGB}{200,200,200}

\newcommand{\method}{\textsc{SemanticKV}}

\newcommand{\fullkvcell}[1]{\cellcolor{softgray}#1}

\providecommand{\planentity}[1]{%
    \textcolor{promptTealBlue!80!black}{#1}%
}

\providecommand{\planaction}[1]{%
    \textcolor{promptPeriwinkle!80!black}{#1}%
}

\providecommand{\plansetting}[1]{%
    \textcolor{promptForestGreen!75!black}{#1}%
}

\providecommand{\planavoid}[1]{%
    \textcolor{promptMelon!55!black}{#1}%
}

\tcbset{
    promptboxbase/.style={
        enhanced,
        breakable,
        sharp corners,
        colback=white,
        coltitle=black,
        boxrule=3pt,
        boxsep=0.5pt,
        shadow={3pt}{-3pt}{0pt}{opacity=1,promptshadow},
        fonttitle=\bfseries,
        listing only,
        listing engine=listings,
        listing options={
            basicstyle=\ttfamily\footnotesize,
            breaklines=true,
            breakatwhitespace=true,
            columns=fullflexible,
            keepspaces=true,
            showstringspaces=false,
            aboveskip=0pt,
            belowskip=0pt
        }
    }
}

\title{Semantically Aligned KV Cache Compression for Efficient Agentic Autoregressive Video Generation}

\newtheorem{skvtheorytheorem}{Theorem}[section]

\author{
    Weizhong Huang\quad Huaijia Kou\quad Tianyu Xie\\
    Xiawu Zheng\quad Liujuan Cao\textsuperscript{*}\\[0.5em]
    \parbox{0.94\textwidth}{\centering\normalsize
    Key Laboratory of Multimedia Trusted Perception and Efficient Computing,\\
    Ministry of Education of China, Xiamen University,\\
    361005, P.R. China.}\\[0.3em]
    {\small\textsuperscript{*}Corresponding author}
}
\date{}

\begin{document}

\maketitle

\begin{abstract}
Agentic autoregressive video generation offers a promising approach to controllable video synthesis through semantic planning, sequential generation, and visual verification. However, growing visual histories enlarge the key--value (KV) cache, increasing memory use and attention costs across candidate rollouts. KV cache compression reduces this overhead, but existing methods primarily optimize individual generators without explicitly accounting for the planner's semantic constraints. Applying them directly can assign low importance to states associated with particular requested attributes or actions, potentially weakening semantic fidelity. In this paper, we introduce \method{}, a training-free method for semantically aligned KV cache compression. To our knowledge, \method{} is the first KV cache compression method for agentic autoregressive video generation that explicitly uses planner-extracted semantic constraints to guide retention. Text cross-attention identifies constraint-related video positions, whose self-attention queries are collected into separate groups. Attention weights are averaged within each group and maximized across groups, giving each cached entry a high score when it is strongly used by any represented constraint. A single top-$k$ selection retains high-scoring entries under the memory budget, leaving their keys, values, and positions unchanged. For fixed query groups and equal entry costs, this selection minimizes an additive upper bound on worst-group discarded attention. The method operates independently within each candidate without retraining or additional video generations. Our evaluation spans four AR video backbones and three benchmarks, covering 5-, 30-, and 60-second generation; our method achieves strong performance across these settings.
On Causal Forcing at a compression ratio of $0.7$, \method{} obtains a VBench Total Score of $84.10$, compared with $84.05$ for the strongest compressed baseline and $83.99$ for Full KV. Meanwhile, it achieves a speedup of $1.22\times$ over Full KV and reduces KV-cache memory from $6.04$ GB to $1.81$ GB, corresponding to a $3.35\times$ reduction.
\end{abstract}

\section{Introduction}

Agentic autoregressive (AR) video generation offers a promising approach to producing controllable videos from complex user requests~\citep{zeng2025coagent,feng2026newton,huang2026vimax}. A semantic planner clarifies the requested content and refines the prompt~\citep{yang2026spiral,cudlenco2026authoring,huang2025planning}, an AR generator synthesizes frames or latent chunks sequentially from previously generated content~\citep{henschel2025streamingt2v,gao2024vid}, and a visual critic assesses generated candidates against the request~\citep{long2026vista}. These complementary roles address the need to both specify intended content and check whether the resulting video follows it. This workflow, however, incurs substantial inference overhead. Each rollout stores historical keys and values in a KV cache to reuse information from earlier chunks~\citep{xing2024live2diff,gao2025ca2}. Longer visual histories increase cache storage and attention computation, while generating multiple candidates repeats these costs even when their caches are not resident simultaneously.

KV cache compression alleviates this bottleneck by reducing the historical states stored and accessed during attention. Representative methods include \textbf{R-KV}~\citep{rkv}, which selects cached states using importance and redundancy; \textbf{Dummy Forcing}~\citep{guo2026efficient}, which allocates context according to attention-head locality; and \textbf{Forcing-KV}~\citep{ji2026forcing}, which applies different retention policies to static and dynamic heads. Other approaches exploit temporal redundancy~\citep{chen2026past}, future-query statistics~\citep{luo2026future}, historical coverage~\citep{zhao2026densitykv}, or instance-specific context representations~\citep{fu2026ispa}. These methods primarily optimize \emph{context use within an individual generator}, rather than explicitly connecting cache retention to the semantic constraints produced by an agentic planner.

Directly applying such policies within an agentic workflow leaves two closely related limitations. First, the planner distinguishes user-specified subjects, attributes, relations, and actions, but cache selection need not identify which states are used by computations associated with each constraint. A detailed plan can therefore guide prompt construction without guiding the history retained during generation. Second, when attention observations are pooled into a global importance score, a state strongly used for one attribute or action can be assigned a low average score because unrelated queries give it little attention. Discarding such states may weaken the information supporting the requested content. The resulting deviations can propagate through subsequent chunks~\citep{liu2026rolling,guo2025end} and compromise long-horizon consistency~\citep{xie2025progressive,wang2025error}. The missing capability is thus not merely stronger compression, but \emph{semantically aligned retention that carries the planner's explicit constraints into cache selection}.

In this paper, we introduce \method{}, a training-free method for semantically aligned KV cache compression in agentic autoregressive video generation. To our knowledge, \method{} is the first KV cache compression method in this setting to explicitly use planner-extracted semantic constraints to guide retention. Its central idea is that \emph{the semantics that guide video generation should also guide the history its generator retains}. The existing semantic constraints $\mathcal{R}$ determine which generator queries are observed and how their cache-attention weights are grouped. This connects retention to the requested content while preventing low scores from unrelated groups from diluting an entry's importance to a particular constraint. The planner, generator, and visual-verification workflow remain unchanged.

Specifically, \method{} first matches each semantic constraint to its words in the candidate prompt. Text cross-attention identifies video positions that attend strongly to these words, and their actual self-attention queries form a constraint-specific group. For each KV entry, the method averages attention weights within each group and takes the maximum across groups to obtain one semantic importance score. Thus, an entry can receive high priority even when it is primarily used by only one represented constraint. At each completed-chunk cache update, a single top-$k$ selection retains the highest-scoring eligible entries within the byte budget, alongside states required by the backbone. Retained keys, values, and positions are packed unchanged, and each candidate uses its own observations and cache. Our local analysis relates discarded attention to groupwise attention-output changes and shows that, for fixed query groups and equal entry costs, top-$k$ selection minimizes an additive upper bound on worst-group discarded attention (Appendix~\ref{app:semantic_theory}). This provides a theoretical basis for the scoring rule without adding another optimization stage.

To evaluate \method{}, we conduct experiments on four AR video backbones: LongLive~\citep{yang2026longlive}, LongLive 2.0~\citep{chen2026longlive}, Self Forcing~\citep{huang2026self}, and Causal Forcing~\citep{zhu2026causal}. We use three benchmarks---VBench~\citep{huang2024vbench}, VBenchLong~\citep{huang2025vbench++}, and VisionReward~\citep{xu2026visionreward}---covering short-video quality, long-horizon consistency, and preference-based assessment at 5-, 30-, and 60-second durations. Comparisons include three KV cache compression baselines and Full KV under matched generation settings and realized cache budgets. Two targeted ablations isolate constraint-conditioned query selection and groupwise score aggregation, while efficiency analysis reports cache memory and generator throughput.
On Causal Forcing at a compression ratio of $0.7$, \method{} obtains a VBench Total Score of $84.10$, compared with $84.05$ for the strongest compressed baseline and $83.99$ for Full KV. Meanwhile, it achieves a speedup of $1.22\times$ over Full KV and reduces KV-cache memory from $6.04$ GB to $1.81$ GB, corresponding to a $3.35\times$ reduction.

\section{Related Work}

\paragraph{Agentic autoregressive video generation.}
Autoregressive video models generate causally ordered frames or latent chunks for streaming and long-horizon synthesis~\citep{chen2024diffusion,chen2025skyreels,teng2025magi,deng2025autoregressive,wang2024loong}, with advances in rollout stability, efficiency, and memory modeling~\citep{liu2026rolling,guo2025end,lin2026autoregressive,yu2025videossm,luo2026quota}. Agentic systems add planning, generation, and feedback~\citep{yuan2024mora,liang2025univa,lin2024videodirectorgpt,xie2024dreamfactory,hu2024storyagent} through staged production, role-based collaboration, and hierarchical planning~\citep{zhuang2024vlogger,xu2025filmagent,wu2025automated}, as well as experience- or feedback-driven refinement~\citep{wang2024aesopagent,soni2024videoagent}. Longer visual histories increase KV-cache costs within each rollout, and candidate generation repeats this overhead. \method{} keeps the planning--generation--verification workflow and AR backbone unchanged, using the planner's semantic constraints to guide cache retention within each candidate.

\paragraph{KV cache compression.}
KV cache compression is a common technique for model compression and acceleration~\citep{huang2025losa,huang2025layerwise,huang2025shapleymoe,zhang2026spt}. KV cache compression reduces inference costs by selecting or compacting cached keys and values. Approaches include merging and salience-based retention~\citep{samuel2026fastar,chen2026past}, compact representations and structured selection~\citep{xi2026quant,ma2026flow,mao2026packforcing,wang2026precisecache,luo2026future}, context management~\citep{li2026packcache,meng2026tethercache,huang2026commitkv,zhang2026kvpo}, and response-aware compression~\citep{feng2025criticalkv,gu2025obcache,zweiger2026attentionmatching,li2026momentkv,yang2026jacap}. Representative methods include R-KV~\citep{rkv}, which selects states by importance and redundancy; Dummy Forcing~\citep{guo2026efficient}, which limits context by head locality; and Forcing-KV~\citep{ji2026forcing}, which preserves transition anchors for static heads and prunes similar segments for dynamic heads. These methods do not explicitly use planner-extracted semantic constraints, potentially overlooking task-relevant states. \method{} instead uses these constraints to identify and retain relevant KV states within the memory budget.

\section{Problem Formulation}
\label{sec:problem_formulation}

\subsection{Agentic AR Video Generation}

Autoregressive (AR) video diffusion models generate video chunks sequentially from a text prompt and previously generated content~\citep{yin2025slow,huang2026self,yang2026longlive}. Given prompt $\mathbf{q}$, the video $\mathbf{X}_{1:T}=(\mathbf{X}_1,\ldots,\mathbf{X}_T)$ is generated as
\begin{equation}
\mathbf{X}_t=\mathcal{G}_{\boldsymbol{\theta}}\left(\mathbf{K}_{t-1},\mathbf{V}_{t-1},\mathbf{q};\boldsymbol{\rho}_t\right),
\label{eq:ar_step}
\end{equation}
where $\mathcal{G}_{\boldsymbol{\theta}}$ is the pretrained generator, $(\mathbf{K}_{t-1},\mathbf{V}_{t-1})$ is the historical KV cache, and $\boldsymbol{\rho}_t$ denotes generation randomness. Although this formulation supports streaming and long-horizon generation~\citep{yin2025slow,huang2026self,yang2026longlive}, early errors can propagate through the generated history~\citep{liu2026rolling,li2026stable}. We therefore adopt a planning--generation--verification workflow~\citep{huang2026genmac,zeng2025coagent}:
\begin{equation}
\mathcal{A}=\left(\mathcal{P}_{\boldsymbol{\phi}},\mathcal{G}_{\boldsymbol{\theta}},\mathcal{C}_{\boldsymbol{\psi}},\pi_{\delta,\tau}\right),
\label{eq:agent}
\end{equation}
where $\mathcal{P}_{\boldsymbol{\phi}}$ is the semantic planner, $\mathcal{G}_{\boldsymbol{\theta}}$ generates candidate videos, $\mathcal{C}_{\boldsymbol{\psi}}$ is the vision--language critic, and $\pi_{\delta,\tau}$ performs conservative candidate selection.

\subsection{Conservative Semantic Planning}
\label{sec:semantic_planning}

The planner refines the input while preserving the original request~\citep{cheng2025vpo,yang2026scmapr}:
\begin{equation}
\left(\mathcal{R},\mathcal{N},\mathbf{q}^{1}\right)=\mathcal{P}_{\boldsymbol{\phi}}(\mathbf{q}),
\qquad \mathbf{q}^{0}=\mathbf{q},
\label{eq:semantic_planning}
\end{equation}
where $\mathcal{R}$ contains semantic constraints such as subjects, attributes, relations, actions, and motion, and $\mathcal{N}$ contains contradictions or unrequested content to avoid. The refined prompt $\mathbf{q}^{1}$ organizes the information in $\mathcal{R}$ while respecting $\mathcal{N}$, whereas $\mathbf{q}^{0}$ keeps the original prompt unchanged. If planning or validation fails, we set $\mathbf{q}^{1}=\mathbf{q}^{0}$. More details are provided in Appendix~\ref{sec:semantic_plan_fields}.

\subsection{Dual-Candidate Autoregressive Generation}
\label{sec:dual_candidate_generation}

Using $\mathbf{q}^{0}$ and $\mathbf{q}^{1}$, the same generator produces two candidates:
\begin{equation}
\mathcal{V}(\mathbf{q})=\left\{\mathbf{X}^{0}_{1:T},\mathbf{X}^{1}_{1:T}\right\},
\qquad
\mathbf{X}^{i}_{t}=\mathcal{G}_{\boldsymbol{\theta}}\left(\mathbf{K}^{i}_{t-1},\mathbf{V}^{i}_{t-1},\mathbf{q}^{i};\boldsymbol{\rho}_{i,t}\right),
\quad i\in\{0,1\},
\label{eq:dual_generation}
\end{equation}
where $i=0$ and $i=1$ denote the original-prompt and planned candidates, respectively. Each candidate maintains an independent KV cache and random-input sequence, and no cached states are shared between them. The original-prompt candidate serves as a fallback~\citep{long2026vista,liu2025video}.

\subsection{Blind Pairwise Visual Verification}
\label{sec:visual_verification}

After generation, the critic compares both candidates against the original request $\mathbf{q}$ and semantic constraints $\mathcal{R}$~\citep{long2026vista}. Their display order is randomized and anonymized to reduce order bias~\citep{zheng2023judging}. The visual input is
\begin{equation}
\mathbf{O}=\mathcal{U}_{m}\left(\mathbf{X}^{\kappa(0)}_{1:T},\mathbf{X}^{\kappa(1)}_{1:T}\right),
\label{eq:visual_observation}
\end{equation}
where $\mathcal{U}_m$ samples $m$ evenly spaced frames from each candidate and $\kappa$ denotes the randomized display order.

The critic evaluates visual quality, semantic fidelity, temporal consistency, and motion correctness~\citep{he2024videoscore}. The utility of candidate $i$ is
\begin{equation}
u_i=w_{\mathrm{v}}z_{\mathrm{v}}^{i}+w_{\mathrm{s}}z_{\mathrm{s}}^{i}+w_{\mathrm{t}}z_{\mathrm{t}}^{i}+w_{\mathrm{m}}z_{\mathrm{m}}^{i},
\label{eq:candidate_utility}
\end{equation}
where the nonnegative weights sum to one. The critic also outputs confidence $\gamma\in[0,1]$, which is used together with $u_0$ and $u_1$ for candidate selection.

\begin{figure}[t]
    \centering
    \includegraphics[width=\linewidth]{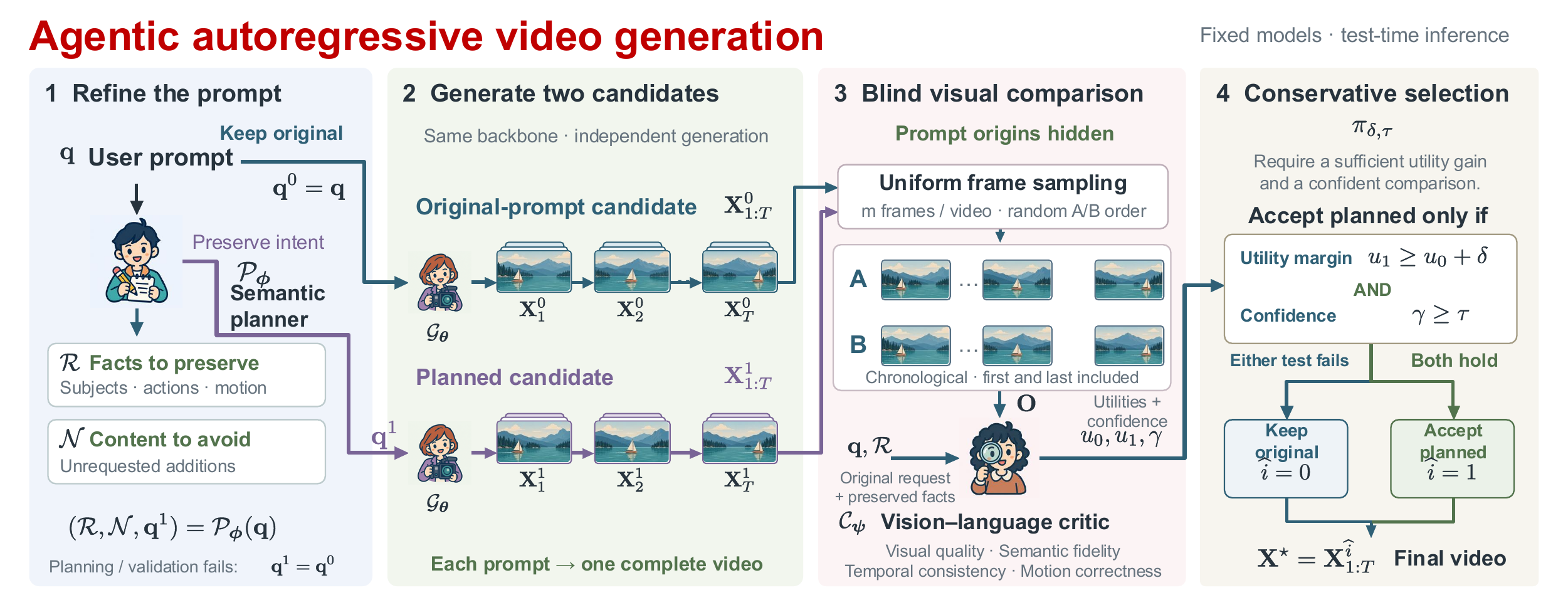}
    \caption{Agentic AR video generation. Semantic constraints $\mathcal{R}$ and prompt-specific exclusions $\mathcal{N}$ guide prompt refinement. The original-prompt and planned candidates from the same generator undergo randomized, anonymous visual verification, followed by conservative selection using Eq.~\ref{eq:selection_rule}.}
    \label{fig:agentic_ar_video_generation}
\end{figure}

\subsection{Conservative Candidate Selection}
\label{sec:candidate_selection}

The planned candidate is selected only when both the utility gain and confidence exceed their thresholds~\citep{zheng2023judging}:
\begin{equation}
\widehat{i}=\pi_{\delta,\tau}(u_0,u_1,\gamma)=
\begin{cases}
1, & u_1 \geq u_0+\delta \ \land\ \gamma\geq\tau,\\
0, & \text{otherwise},
\end{cases}
\label{eq:selection_rule}
\end{equation}
where $\delta>0$ is the minimum utility gain and $\tau$ is the confidence threshold. The final output is
\begin{equation}
\mathbf{X}^{\star}=\mathbf{X}^{\widehat{i}}_{1:T}.
\label{eq:final_output}
\end{equation}
Figure~\ref{fig:agentic_ar_video_generation} summarizes the complete planning--generation--verification workflow.

\subsection{KV Cache Compression Problem}
\label{sec:kv_cache_problem}

We compress each candidate's KV cache during AR generation~\citep{ji2026forcing,guo2026efficient} without changing planning, verification, or selection. Omitting candidate indices, the compression policy is
\begin{equation}
(\widetilde{\mathbf{K}}_t,\widetilde{\mathbf{V}}_t)=\mathcal{C}_{\eta}(\mathbf{K}_t,\mathbf{V}_t;\mathcal{R},\mathbf{q}^{i}),
\label{eq:kv_cache_compression}
\end{equation}
where $\eta$ denotes the policy settings. The next chunk is generated using the compressed history:
\begin{equation}
\widetilde{\mathbf{X}}_{t+1}^{i}=\mathcal{G}_{\boldsymbol{\theta}}\left(\widetilde{\mathbf{K}}_t,\widetilde{\mathbf{V}}_t,\mathbf{q}^{i};\boldsymbol{\rho}_{i,t+1}\right).
\label{eq:compressed_ar_step}
\end{equation}

Let $(\mathbf{K}^{\mathrm{full}}_t,\mathbf{V}^{\mathrm{full}}_t)$ denote the uncompressed reference under the same native schedule. We aim to maximize video quality under a bounded cache budget:
\begin{equation}
\max_{\eta}\;\mathbb{E}\!\left[\mathcal{Q}\!\left(\widetilde{\mathbf{X}}_{1:T},\mathbf{q}\right)\right]
\quad\text{s.t.}\quad
\max_t\frac{M(\widetilde{\mathbf{K}}_t,\widetilde{\mathbf{V}}_t)}{M(\mathbf{K}^{\mathrm{full}}_t,\mathbf{V}^{\mathrm{full}}_t)}\leq 1-r,
\quad 0\leq r<1,
\label{eq:kv_cache_objective}
\end{equation}
where $\mathcal{Q}$ measures video quality, $r$ is the compression ratio (the fraction of cache removed), and $M$ measures KV cache storage.

\section{Methodology}
\label{sec:method}

\method{} implements the compression policy $\mathcal{C}_{\eta}$ in Eq.~\ref{eq:kv_cache_compression} by retaining cached information used by queries associated with the planner's semantic constraints $\mathcal{R}$. We first identify these queries, then aggregate their attention into one score per KV entry and apply a single top-$k$ selection under the memory constraint in Eq.~\ref{eq:kv_cache_objective}.

\subsection{Constraint-Conditioned Query Selection}
\label{sec:constraint_probes}

\paragraph{Connecting semantic constraints to video positions.}
For each layer and KV head, let $\mathcal{H}_t$ index the available entries in $(\mathbf{K}_t,\mathbf{V}_t)$ after chunk $t$ of candidate $i$, with $\mathcal{A}_t\subseteq\mathcal{H}_t$ containing entries required by the backbone. We match the content words of each semantic constraint $r_a\in\mathcal{R}$ to positions $\mathcal{T}_a\subseteq\mathcal{T}_{\mathrm{text}}$ in the candidate prompt $\mathbf{q}^{i}$, where $\mathcal{T}_{\mathrm{text}}$ contains the text positions visible under the native cross-attention mask.

For video-token position $x$ in the completed chunk and text position $b\in\mathcal{T}_{\mathrm{text}}$, let $\mathbf{z}^{\mathrm{cross}}_{x,h}$ and $\mathbf{k}^{\mathrm{text}}_{b,h}$ denote the cross-attention query and text key in routing head $h$, including native normalization. With $H_{\mathrm{c}}$ routing heads and query/key dimension $d_{\mathrm{c}}$, we compute
\begin{equation}
\alpha^{\mathrm{text}}_{xb}=\frac{1}{H_{\mathrm{c}}}\sum_{h=1}^{H_{\mathrm{c}}}\frac{\exp\!\left((\mathbf{z}^{\mathrm{cross}}_{x,h})^{\top}\mathbf{k}^{\mathrm{text}}_{b,h}/\sqrt{d_{\mathrm{c}}}\right)}{\sum_{b'\in\mathcal{T}_{\mathrm{text}}}\exp\!\left((\mathbf{z}^{\mathrm{cross}}_{x,h})^{\top}\mathbf{k}^{\mathrm{text}}_{b',h}/\sqrt{d_{\mathrm{c}}}\right)},
\qquad
g_a(x)=\sum_{b\in\mathcal{T}_a}\alpha^{\mathrm{text}}_{xb}.
\label{eq:atom_routing}
\end{equation}
Here, $\alpha^{\mathrm{text}}_{xb}$ is the head-averaged attention from video position $x$ to text token $b$. A larger routing score $g_a(x)$ indicates stronger attention to the words matched to constraint $r_a$.

\paragraph{Collecting the corresponding queries.}
For each reliably matched constraint, we select up to $n_q$ positions with the largest $g_a(x)$, where $n_q$ is the per-group query limit. Let $\mathbf{z}_p$ denote an existing self-attention query collected at a selected position, with $p$ indexing the query observation, and let $\mathcal{P}_a$ contain the query indices for constraint $r_a$. Unlike $\mathbf{z}^{\mathrm{cross}}_{x,h}$, which attends to text keys, $\mathbf{z}_p$ attends to video keys in $\mathbf{K}_t$. Thus, cross-attention selects where to collect queries, while self-attention measures which cache entries those positions use.

\subsection{Semantically Aligned KV Retention}
\label{sec:budgeted_selection}

\paragraph{From selected queries to cache-attention weights.}
Let $\mathcal{I}\neq\varnothing$ index the nonempty query groups used for scoring. For entry $j\in\mathcal{H}_t$, let $\mathbf{k}_j$ and $\mathbf{v}_j$ denote its key and value, and let $\mathcal{H}_{t,p}\subseteq\mathcal{H}_t$ contain the entries visible to query $\mathbf{z}_p$ under the native attention mask. With query/key dimension $d_k$, we compute
\begin{equation}
\alpha_{pj}=\frac{\exp(\mathbf{z}_p^{\top}\mathbf{k}_j/\sqrt{d_k})}{\sum_{j'\in\mathcal{H}_{t,p}}\exp(\mathbf{z}_p^{\top}\mathbf{k}_{j'}/\sqrt{d_k})},
\quad j\in\mathcal{H}_{t,p},
\qquad \alpha_{pj}=0\ \text{otherwise}.
\label{eq:semantic_attention}
\end{equation}
Unlike $\alpha^{\mathrm{text}}_{xb}$ in Eq.~\ref{eq:atom_routing}, these weights describe attention to cached video entries rather than prompt tokens. We average them within each group and take the maximum across groups:
\begin{equation}
w_{aj}=\frac{1}{|\mathcal{P}_a|}\sum_{p\in\mathcal{P}_a}\alpha_{pj},
\qquad
s_j=\max_{a\in\mathcal{I}}w_{aj}.
\label{eq:semantic_score}
\end{equation}
Here, $w_{aj}$ is group $a$'s mean attention to entry $j$, and $s_j$ is its semantic importance. The mean avoids favoring a group merely because it contains more queries; the maximum gives an entry high priority when strongly used by any represented constraint.

\paragraph{Why discarded attention matters.}
For a retained set $\mathcal{A}_t\subseteq\mathcal{S}\subseteq\mathcal{H}_t$, define the average attention discarded from group $a$ and the largest such loss as
\begin{equation}
\ell_a(\mathcal{S})=\sum_{j\in\mathcal{H}_t\setminus\mathcal{S}}w_{aj},
\qquad
\mathcal{L}(\mathcal{S})=\max_{a\in\mathcal{I}}\ell_a(\mathcal{S}).
\label{eq:theory_group_mass}
\end{equation}
We compare attention before and after retention with the observed queries, retained keys, values, positions, and native masks held fixed. Let $\mathbf{o}_p$ and $\mathbf{o}_p^{\mathcal{S}}$ denote the corresponding value-weighted outputs before the output projection; retained attention is renormalized over the remaining visible entries.

\begin{skvtheorytheorem}[Discarded attention controls local output error]
\label{thm:semantic_mass_error}
\label{lem:semantic_mass_error}
Suppose $\mathcal{S}$ retains at least one visible entry for every selected query. Let $D_{\mathrm{V}}=\max_{j,j'\in\mathcal{H}_t}\|\mathbf{v}_j-\mathbf{v}_{j'}\|_2$ be the largest distance between available value vectors. Then the largest group-average attention-output error satisfies
\begin{equation}
\mathcal{E}(\mathcal{S};\mathbf{V}_t)
:=\max_{a\in\mathcal{I}}\frac{1}{|\mathcal{P}_a|}\sum_{p\in\mathcal{P}_a}\|\mathbf{o}_p-\mathbf{o}_p^{\mathcal{S}}\|_2
\leq D_{\mathrm{V}}\,\mathcal{L}(\mathcal{S}).
\label{eq:theory_group_error}
\end{equation}
\end{skvtheorytheorem}

Theorem~\ref{thm:semantic_mass_error} connects attention retention to a local output guarantee: limiting the discarded mass in every observed group limits its average output change. The result includes attention renormalization and does not require the discarded mass to be small. Its worst-case sharpness is established in Appendix~\ref{app:semantic_error_bound}.

\paragraph{Why use the groupwise maximum.}
Directly minimizing $\mathcal{L}$ couples the retention decisions because its worst-affected group can change with the retained set. The score $s_j$ instead gives each entry a fixed contribution to an upper bound.

\begin{skvtheorytheorem}[Smallest universal additive bound]
\label{thm:semantic_additive_envelope}
For every $\mathcal{A}_t\subseteq\mathcal{S}\subseteq\mathcal{H}_t$,
\begin{equation}
\mathcal{L}(\mathcal{S})\leq\mathcal{U}(\mathcal{S})
:=\sum_{j\in\mathcal{H}_t\setminus\mathcal{S}}s_j
\leq |\mathcal{I}|\,\mathcal{L}(\mathcal{S}).
\label{eq:theory_envelope_sandwich}
\end{equation}
Among bounds of the form $\sum_{j\in\mathcal{H}_t\setminus\mathcal{S}}b_j$ with fixed nonnegative coefficients that hold for every deletion of optional entries, the smallest valid coefficients are $b_j=s_j$ for $j\in\mathcal{H}_t\setminus\mathcal{A}_t$.
\end{skvtheorytheorem}

Thus, the maximum is not an arbitrary aggregation: it gives the smallest entrywise penalties that upper-bound every group's discarded attention for all deletion sets. In contrast, pooling queries controls a query-count-weighted average and can dilute an entry's importance to a particular constraint (Appendix~\ref{app:semantic_grouping_theory}). This does not imply that every constraint receives reserved memory or is satisfied in the generated video.

\begin{figure}[t]
    \centering
    \includegraphics[width=\linewidth]{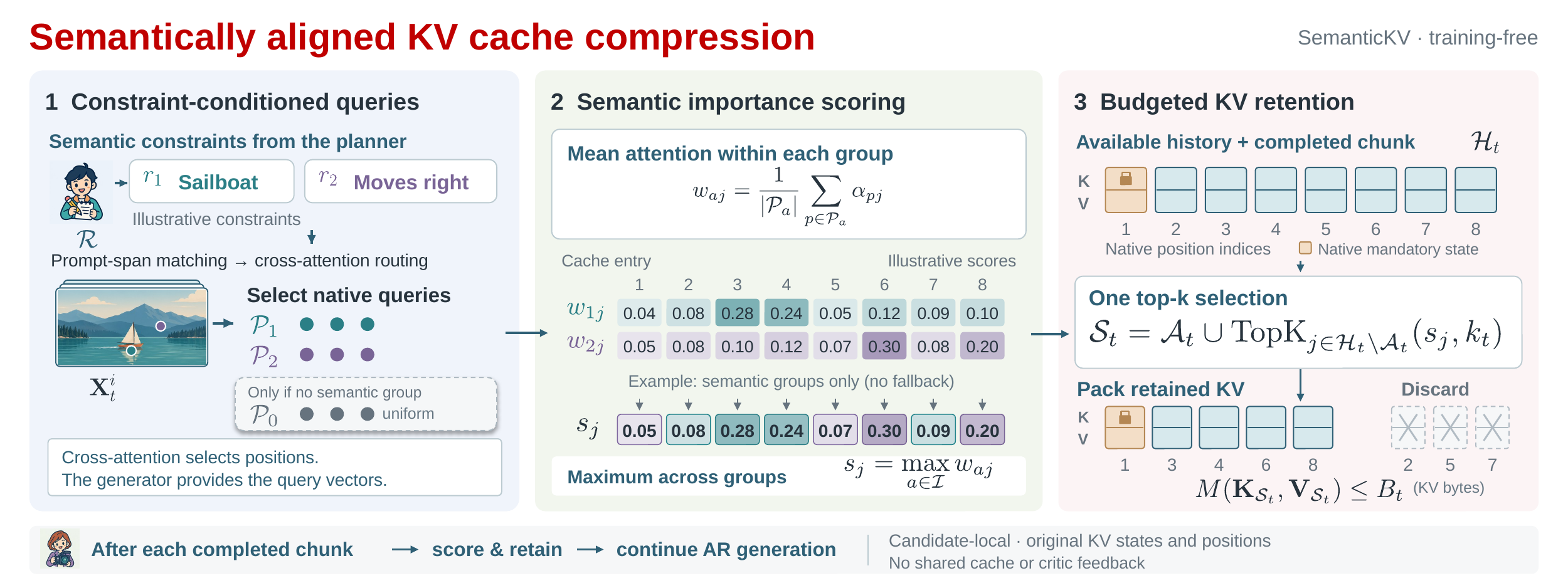}
    \caption{Overview of \method{}. Semantic constraints $\mathcal{R}$ are matched to prompt tokens. Text cross-attention identifies related video positions, from which we collect the generator's self-attention queries. Their attention to each KV entry is averaged within groups and maximized across groups. A single top-$k$ selection retains high-scoring entries within the byte budget, together with states required by the backbone. Retained keys, values, and positions remain unchanged.}
    \label{fig:semantickv_methodology}
\end{figure}

\paragraph{From the memory constraint to retained entries.}
We allocate the total byte budget implied by compression ratio $r$ in Eq.~\ref{eq:kv_cache_objective} across layer/KV-head units in proportion to their uncompressed reference sizes. Let $B_t$ denote the current unit's budget. Mandatory entries $\mathcal{A}_t$ count toward $B_t$; \method{} adds no further protected states. Since entries within a unit have equal native-precision KV costs, let $k_t$ be the largest number of additional available entries that fit after accounting for mandatory states, position records, metadata, and allocation granularity. We retain
\begin{equation}
\mathcal{S}_t=\mathcal{A}_t\cup
\operatorname{TopK}_{j\in\mathcal{H}_t\setminus\mathcal{A}_t}(s_j,k_t),
\qquad M(\mathbf{K}_{\mathcal{S}_t},\mathbf{V}_{\mathcal{S}_t})\leq B_t.
\label{eq:semantic_retention}
\end{equation}
Here, $\operatorname{TopK}$ returns the $k_t$ highest-scoring eligible indices. The selected keys and values form the compressed cache in Eq.~\ref{eq:kv_cache_compression}; $M$ includes persistent position records and metadata. Budgets that cannot hold the mandatory states and a usable attention context are reported as infeasible.

\begin{skvtheorytheorem}[Top-$k$ optimality for the bound]
\label{thm:semantic_topk_bound}
Suppose optional entries have equal storage costs and budget feasibility depends only on their count. Let $\mathcal{F}_t$ contain all retained sets that include $\mathcal{A}_t$ and at most $k_t$ optional entries. The set $\mathcal{S}_t$ in Eq.~\ref{eq:semantic_retention} satisfies
\begin{equation}
\mathcal{U}(\mathcal{S}_t)=\min_{\mathcal{S}\in\mathcal{F}_t}\mathcal{U}(\mathcal{S}),
\qquad
\mathcal{L}(\mathcal{S}_t)\leq |\mathcal{I}|\min_{\mathcal{S}\in\mathcal{F}_t}\mathcal{L}(\mathcal{S}).
\label{eq:theory_topk_optimality}
\end{equation}
\end{skvtheorytheorem}

Theorem~\ref{thm:semantic_topk_bound} explains why one ranking is sufficient to minimize the additive bound. The second inequality compares its worst-group discarded mass with the best possible value under the same count budget. For a usable $\mathcal{S}_t$, Theorem~\ref{thm:semantic_mass_error} also gives $\mathcal{E}(\mathcal{S}_t;\mathbf{V}_t)\leq D_{\mathrm{V}}\min_{\mathcal{S}\in\mathcal{F}_t}\mathcal{U}(\mathcal{S})$. These are guarantees for fixed observed queries, not exact minimization of actual output error or final video-quality loss.

\paragraph{Returning to autoregressive generation.}
If all available entries fit, the cache remains unchanged. Otherwise, we score once, select once, and pack the retained keys, values, and position records in their original logical order, breaking score ties by original position index. Retained vectors and positions remain unchanged, with no rescoring after deletion. The compressed cache generates the next chunk as in Eq.~\ref{eq:compressed_ar_step}; subsequent updates consider retained history together with new KV states. Both candidates use the same policy and fixed $\mathcal{R}$, but their own prompts, queries, and caches, without critic feedback during generation. Query collection, scoring, selection, and packing count toward runtime, and their temporary buffers count toward peak memory. Figure~\ref{fig:semantickv_methodology} illustrates the complete procedure, and Algorithm~\ref{alg:SemanticKV} summarizes the cache update.

\section{Experiments}
\label{sec:video_experiments}

\subsection{Experimental Setup}

\paragraph{Models.}
We evaluate four autoregressive video diffusion models spanning different model designs and parameter scales: LongLive~\citep{yang2026longlive}, LongLive 2.0~\citep{chen2026longlive}, Self Forcing~\citep{huang2026self}, and Causal Forcing~\citep{zhu2026causal}. The evaluated models cover a broad range of model sizes, with parameter counts ranging from 1.3B to 5B.

\paragraph{Baselines.}
We compare \method{} with representative KV-cache compression methods that provide publicly available implementations and can be reliably integrated into autoregressive video diffusion models. Specifically, we compare with \textbf{Full KV}, \textbf{R-KV}~\citep{rkv}, \textbf{Dummy Forcing}~\citep{guo2026efficient}, and \textbf{Forcing-KV}~\citep{ji2026forcing}. For all compressed methods, we follow their official implementations and apply the same KV-cache compression ratio whenever possible to ensure a fair comparison. Appendix~\ref{sec:Details_of_Baselines} provides further details.

\paragraph{Benchmarks.}
We evaluate video generation quality using three complementary benchmarks: VBench~\citep{huang2024vbench}, VBenchLong~\citep{huang2025vbench++}, and VisionReward~\citep{xu2026visionreward}. For short-video evaluation, we generate 5-second videos using all 946 official VBench prompts. For long-video evaluation, we use 128 prompts from MovieGen~\citep{polyak2024movie} to generate 30-second and 60-second videos, with five random seeds for each prompt. We additionally employ VisionReward on the 100-prompt evaluation set used by Causal Forcing. We evaluate all 16 VBench dimensions. Detailed metric definitions appear in Appendix~\ref{app:video_metrics}.

\paragraph{Implementation Details.}
All experiments are run on a server equipped with 8$\times$ NVIDIA H20 96GB GPUs. Our implementation uses PyTorch for video generation and serves Qwen3.5-9B through vLLM for semantic planning and pairwise visual verification. The default launcher assigns seven GPUs to independent generation workers and one GPU to planning and verification. All four video backbones use BF16 precision and four denoising steps; model-specific resolutions, frame rates, and chunk sizes are detailed in Appendix~\ref{sec:Details_of_Models}. We evaluate five base seeds, 42--46, generating two candidates sequentially per run. The critic samples $m=6$ frames per candidate, and candidate selection uses $\delta=1.0$, $\tau=0.55$, and utility weights $(w_{\mathrm{v}},w_{\mathrm{s}},w_{\mathrm{t}},w_{\mathrm{m}})=(0.4,0.2,0.3,0.1)$. Both the planner and critic use temperature zero.

\begin{table*}[t!]
\centering
\captionsetup{skip=2pt}
\fontsize{8.0}{8.6}\selectfont
\setlength{\tabcolsep}{2.0pt}
\renewcommand{\arraystretch}{1.15}

\caption{Five-second VBench at compression ratio $0.7$: Total Score, temporal quality, and six shared semantic diagnostics. Table~\ref{tab:main_result_backbone} gives all 16 dimensions. Full KV is the uncompressed reference; efficiency is DiT-only. $\uparrow$/$\downarrow$: higher/lower is better. \textbf{Bold}: best compressed result for Total Score.}
\label{tab:main_result_short}

\resizebox{\textwidth}{!}{%
\begin{tabular}{@{}ll*{14}{c}@{}}
\toprule

\multirow{2}{*}{\textbf{Backbone}}
& \multirow{2}{*}{\textbf{Method}}
& \multicolumn{2}{c}{\textbf{Efficiency}}
& \multicolumn{2}{c}{\textbf{KV Cache}}
& \multicolumn{3}{c}{\textbf{Temporal Quality}}
& \multicolumn{7}{c}{\textbf{VBench: Overall and Semantic Fidelity}} \\

\cmidrule(lr){3-4}
\cmidrule(lr){5-6}
\cmidrule(lr){7-9}
\cmidrule(lr){10-16}

&
& \textbf{FPS}$\uparrow$
& \textbf{Speedup}$\uparrow$
& \shortstack{\textbf{Memory}\textbf{ (GB)}$\downarrow$}
& \shortstack{\textbf{Reduction}\textbf{ ($\times$)}$\uparrow$}
& \textbf{Disc.}$\downarrow$
& \textbf{Dynamic}$\uparrow$
& \textbf{Smooth.}$\uparrow$
& \shortstack{\textbf{Total}\\\textbf{Score}$\uparrow$}
& \shortstack{\textbf{Object}\\\textbf{Class}$\uparrow$}
& \shortstack{\textbf{Multiple}\\\textbf{Objects}$\uparrow$}
& \shortstack{\textbf{Human}\\\textbf{Action}$\uparrow$}
& \textbf{Color}$\uparrow$
& \shortstack{\textbf{Spatial}\\\textbf{Relation}$\uparrow$}
& \textbf{Scene}$\uparrow$ \\

\midrule

\multirow{5}{*}{\shortstack{Self\\Forcing}}
& \fullkvcell{Full KV}
& \fullkvcell{6.95}
& \fullkvcell{1.00$\times$}
& \fullkvcell{6.04} & \fullkvcell{1.00$\times$}
& \fullkvcell{2.25}
& \fullkvcell{91.67}
& \fullkvcell{97.77}
& \fullkvcell{82.96}
& \fullkvcell{89.83}
& \fullkvcell{75.55}
& \fullkvcell{92.00}
& \fullkvcell{90.54}
& \fullkvcell{71.67}
& \fullkvcell{48.05} \\

& R-KV
& 7.36
& 1.06$\times$
& 1.81 & 3.35$\times$
& 2.23
& 92.78
& 97.77
& 82.90
& 88.29
& 74.15
& 93.00
& 91.40
& 71.66
& 47.53 \\

& Dummy Forcing
& 8.16
& 1.17$\times$
& 1.81 & 3.35$\times$
& 2.23
& 91.39
& 97.81
& 82.87
& 89.26
& 77.07
& 92.00
& 91.12
& 71.01
& 47.43 \\

& Forcing-KV
& 7.92
& 1.14$\times$
& 1.81 & 3.35$\times$
& 2.62
& 92.78
& 97.55
& 82.69
& 88.21
& 75.88
& 92.60
& 90.64
& 71.44
& 47.72 \\

\rowcolor{softcyan}
& \textbf{\method{}}
& 7.95
& 1.14$\times$
& 1.81 & 3.35$\times$
& 2.26
& 93.33
& 97.80
& \textbf{83.03}
& 89.43
& 75.47
& 92.00
& 91.30
& 72.46
& 48.65 \\

\midrule

\multirow{5}{*}{\shortstack{Causal\\Forcing}}
& \fullkvcell{Full KV}
& \fullkvcell{7.24}
& \fullkvcell{1.00$\times$}
& \fullkvcell{6.04} & \fullkvcell{1.00$\times$}
& \fullkvcell{2.39}
& \fullkvcell{83.61}
& \fullkvcell{97.87}
& \fullkvcell{83.99}
& \fullkvcell{91.11}
& \fullkvcell{75.88}
& \fullkvcell{90.00}
& \fullkvcell{88.72}
& \fullkvcell{77.57}
& \fullkvcell{46.25} \\

& R-KV
& 8.14
& 1.13$\times$
& 1.81 & 3.35$\times$
& 2.54
& 86.39
& 97.72
& 84.00
& 92.88
& 76.77
& 91.40
& 87.89
& 75.60
& 45.20 \\

& Dummy Forcing
& 9.20
& 1.27$\times$
& 1.81 & 3.35$\times$
& 2.53
& 86.39
& 97.78
& 84.05
& 91.72
& 76.34
& 91.60
& 88.12
& 77.57
& 43.88 \\

& Forcing-KV
& 8.91
& 1.23$\times$
& 1.81 & 3.35$\times$
& 2.97
& 90.00
& 97.39
& 84.04
& 92.74
& 79.19
& 89.80
& 89.27
& 78.18
& 43.18 \\

\rowcolor{softcyan}
& \textbf{\method{}}
& 8.83
& 1.22$\times$
& 1.81 & 3.35$\times$
& 2.52
& 86.67
& 97.72
& \textbf{84.10}
& 91.85
& 77.09
& 91.60
& 89.09
& 78.01
& 45.35 \\

\bottomrule
\end{tabular}%
}
\end{table*}

\subsection{Main Results}
\label{sec:video_main_results}

\paragraph{Short-video quality and efficiency.}
Table~\ref{tab:main_result_short} compares 5-second video generation on Self Forcing and Causal Forcing. \method{} achieves the highest Total Score among all compressed methods on both backbones, reaching $83.03$ on Self Forcing and $84.10$ on Causal Forcing, outperforming the strongest compressed baselines by $0.13$ and $0.05$ points, respectively. Meanwhile, \method{} increases DiT-only throughput from $6.95$ to $7.95$ FPS on Self Forcing and from $7.24$ to $8.83$ FPS on Causal Forcing, corresponding to $1.14\times$ and $1.22\times$ speedups over Full KV. It also reduces KV-cache memory from $6.04$ GB to $1.81$ GB, corresponding to a $3.35\times$ reduction. These results demonstrate that \method{} provides a favorable balance between generation quality, memory efficiency, and inference speed. The complete results across all 16 VBench dimensions are reported in Table~\ref{tab:main_result_backbone} in Appendix~\ref{app:full_benchmark_results}.

\begin{figure*}[t]
    \centering
    \includegraphics[width=\textwidth]{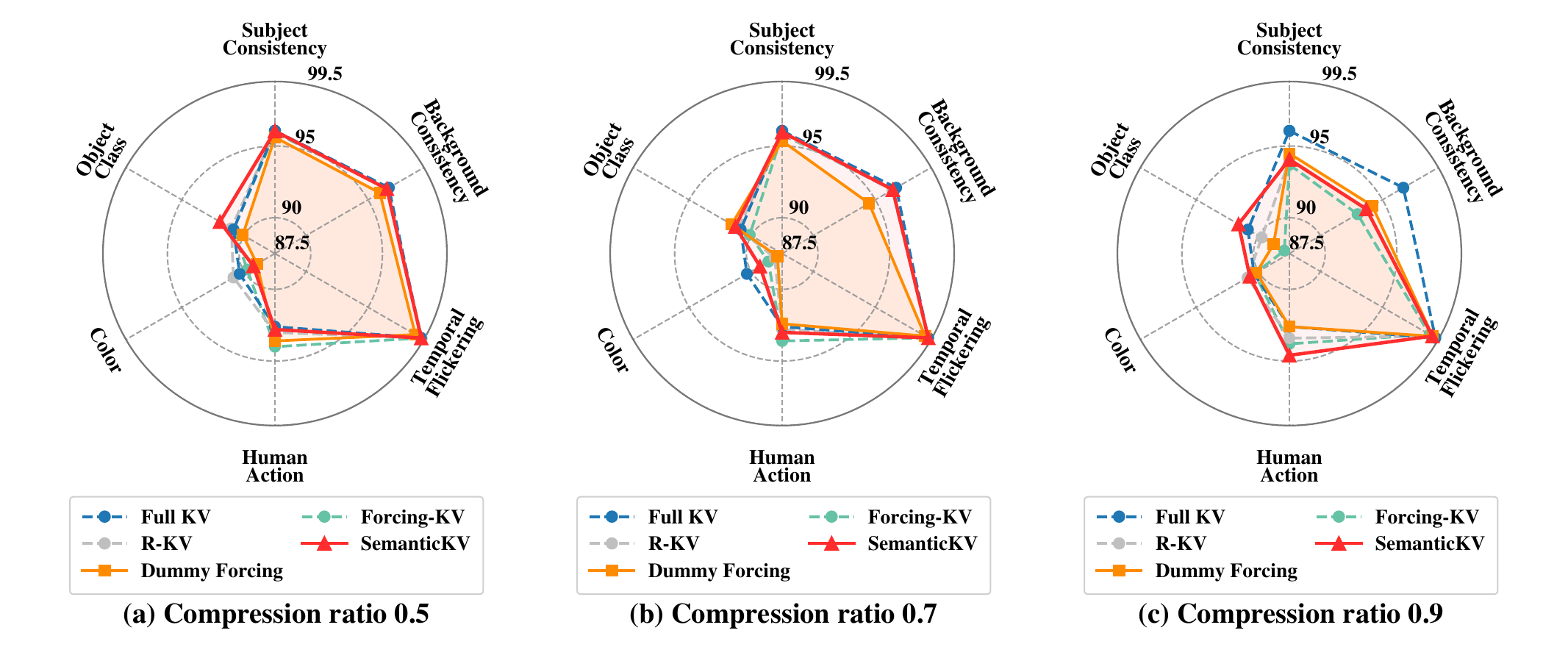}
    \caption{Five-second VBench across compression ratios $0.5/0.7/0.9$, with the same five methods and six dimensions; higher is better. Table~\ref{tab:main_result_short_full} gives exact scores and DiT-only efficiency.}
    \label{fig:cache_setting_radar}
\end{figure*}

\paragraph{Quality across compression ratios.}
Figure~\ref{fig:cache_setting_radar} shows that \method{} maintains strong semantic quality across compression ratios of $0.5$, $0.7$, and $0.9$. It achieves the best compressed result on several semantic dimensions, with especially clear gains at $0.9$, where Object Class improves by up to $3.69$ points over competing methods and Human Action reaches $94.60$. Meanwhile, throughput increases from $8.00$ to $10.00$ FPS, reaching a $1.52\times$ speedup over Full KV at $0.9$. These results demonstrate a favorable quality--efficiency trade-off across different compression ratios.

\paragraph{Long-horizon generation.}
On LongLive 2.0, \method{} achieves $1.27\times$ and $1.29\times$ DiT-only speedups for 30- and 60-second generation, while the Total Score decreases by only $0.78$ and $1.28$ points relative to Full KV, respectively. These results show that \method{} extends its efficiency gains to substantially longer autoregressive generation. Complete VBenchLong results are provided in Appendix~\ref{app:vbenchlong_full}.

\paragraph{Preference-based evaluation.}
On the VisionReward benchmark, \method{} achieves an instruction-following score of $98.000$, matching the best compressed methods and exceeding Full KV and Dummy Forcing by $2.0$ points. Its overall VisionReward score is also higher than R-KV and Forcing-KV. Detailed preference-based results are reported in Appendix~\ref{app:visionreward_full}.

\subsection{Ablation Studies}
\label{sec:video_ablation}

\paragraph{Component ablations.}
We ablate constraint-conditioned query selection and groupwise aggregation under the same cache budget. \method{} improves the Semantic Score from $76.94$ to $77.18$ over uniform queries and from $76.78$ to $77.18$ over pooled constraints, while maintaining similar throughput. These results support the contributions of both semantic query selection and constraint-wise aggregation. Complete results are reported in Appendix~\ref{app:ablation_efficiency}.

\paragraph{Query limit.}
With the cache budget fixed, $n_q=32$ achieves the highest Total Score ($81.68$) among $n_q\in\{16,32,64\}$ while reaching $10.00$ FPS. We therefore use $n_q=32$ as the default setting; detailed results are provided in Appendix~\ref{sec:query_limit_ablation}.

\subsection{Additional Ablation Studies}
\label{sec:additional_video_ablation}

\paragraph{Paired semantic analysis.}
On matched prompt--seed outputs, \method{} shows positive gains in 15 of 18 semantic dimension--baseline comparisons. The six-dimension macro improvements over R-KV, Dummy Forcing, and Forcing-KV are $1.00$, $0.86$, and $1.25$ points, respectively, indicating a consistent advantage in semantic fidelity. Detailed paired results and bootstrap intervals are provided in Appendix~\ref{app:paired_reliability}.

\paragraph{Semantic complexity.}
We further analyze performance with respect to the number of planner-extracted semantic constraints. The gains of \method{} become more pronounced on multi-constraint prompts, reaching macro improvements of $4.62$, $3.26$, and $7.24$ points over R-KV, Dummy Forcing, and Forcing-KV for two-constraint prompts. Results for different semantic-complexity groups are reported in Appendix~\ref{app:semantic_complexity}.

\paragraph{Temporal semantic drift.}
We also evaluate within-video semantic drift using CLIP-based prompt alignment and visual similarity. The compressed methods show only small differences, with all paired confidence intervals between \method{} and the compressed baselines including zero. This suggests that the semantic gains of \method{} do not introduce detectable additional temporal drift on this subset. Detailed results are provided in Appendix~\ref{app:temporal_drift}.

\section{Conclusion}

We introduce \method{}, a training-free method for semantically aligned KV cache compression in agentic autoregressive video generation. Its central idea is to use the planner's semantic constraints not only to guide generation, but also to guide cache retention. Constraint-conditioned queries provide groupwise attention scores, and one top-$k$ selection retains high-scoring KV entries under the byte budget. The method preserves native keys, values, and positions and leaves the outer agentic workflow unchanged. Its practical benefit is assessed through matched semantic-quality, cache-memory, and throughput measurements rather than inferred from attention scores alone.

\section*{Statement on LLM Usage}

LLMs contributed to the research workflow by supporting idea exploration, experimental planning, literature discovery, manuscript drafting and revision, and mathematical reasoning. They were additionally used for language refinement and code-related assistance. All resulting suggestions, derivations, references, implementations, and interpretations were carefully examined and independently validated by the authors. The authors remain solely responsible for the scientific content, methodological choices, reported results, and conclusions of this paper.

\section*{Ethics Statement}

The research presented in this paper does not involve human participants, personally identifiable information, or sensitive private data. We have conducted this work in accordance with the ICLR Code of Ethics and are not aware of any particular ethical risks arising from the proposed method or the reported experiments.

\section*{Reproducibility Statement}

The experimental configuration, including the evaluated models, datasets, baselines, implementation details, and evaluation procedures, is described in the Experiments section. We additionally provide our code as part of the supplementary materials to support reproducibility. The complete implementation will be released publicly after acceptance.

\bibliography{references}

\begin{thebibliography}{80}
\providecommand{\natexlab}[1]{#1}
\providecommand{\url}[1]{\texttt{#1}}
\expandafter\ifx\csname urlstyle\endcsname\relax
  \providecommand{\doi}[1]{doi: #1}\else
  \providecommand{\doi}{doi: \begingroup \urlstyle{rm}\Url}\fi

\bibitem[ai et~al.(2025)ai, Teng, Jia, Sun, Li, Li, Tang, Han, Zhang, Zhang, Luo, Kang, Sun, Cao, Huang, Lin, Fang, Tao, Zhang, Wang, Liu, Shi, Su, Sun, Pan, Wang, Sheng, Cui, Hu, Yan, Yin, Zhang, Liu, Yin, Yang, Song, Hu, Zhang, and Li]{teng2025magi}
Sand. ai, Hansi Teng, Hongyu Jia, Lei Sun, Lingzhi Li, Maolin Li, Mingqiu Tang, Shuai Han, Tianning Zhang, W.~Q. Zhang, Weifeng Luo, Xiaoyang Kang, Yuchen Sun, Yue Cao, Yunpeng Huang, Yutong Lin, Yuxin Fang, Zewei Tao, Zheng Zhang, Zhongshu Wang, Zixun Liu, Dai Shi, Guoli Su, Hanwen Sun, Hong Pan, Jie Wang, Jiexin Sheng, Min Cui, Min Hu, Ming Yan, Shucheng Yin, Siran Zhang, Tingting Liu, Xianping Yin, Xiaoyu Yang, Xin Song, Xuan Hu, Yankai Zhang, and Yuqiao Li.
\newblock Magi-1: Autoregressive video generation at scale, 2025.
\newblock URL \url{https://arxiv.org/abs/2505.13211}.

\bibitem[Cai et~al.(2025)Cai, Xiao, Sun, Luo, Zhang, Wan, Li, Zhou, Chang, Gu, Dong, Anandkumar, Asi, and Hu]{rkv}
Zefan Cai, Wen Xiao, Hanshi Sun, Cheng Luo, Yikai Zhang, Ke~Wan, Yucheng Li, Yeyang Zhou, Li-Wen Chang, Jiuxiang Gu, Zhen Dong, Animashree Anandkumar, Abedelkadir Asi, and Junjie Hu.
\newblock R-kv: Redundancy-aware kv cache compression for reasoning models.
\newblock \emph{Advances in Neural Information Processing Systems}, 38:\penalty0 60980--61005, 2025.

\bibitem[Chen et~al.(2024)Chen, Mart{\'\i}~Mons{\'o}, Du, Simchowitz, Tedrake, and Sitzmann]{chen2024diffusion}
Boyuan Chen, Diego Mart{\'\i}~Mons{\'o}, Yilun Du, Max Simchowitz, Russ Tedrake, and Vincent Sitzmann.
\newblock Diffusion forcing: Next-token prediction meets full-sequence diffusion.
\newblock \emph{Advances in Neural Information Processing Systems}, 37:\penalty0 24081--24125, 2024.

\bibitem[Chen et~al.(2025)Chen, Lin, Yang, Lin, Zhu, Fan, Zhang, Chen, Chen, Ma, et~al.]{chen2025skyreels}
Guibin Chen, Dixuan Lin, Jiangping Yang, Chunze Lin, Junchen Zhu, Mingyuan Fan, Hao Zhang, Sheng Chen, Zheng Chen, Chengcheng Ma, et~al.
\newblock Skyreels-v2: Infinite-length film generative model.
\newblock \emph{arXiv preprint arXiv:2504.13074}, 2025.

\bibitem[Chen et~al.(2026{\natexlab{a}})Chen, Xu, Yang, Chen, and Deng]{chen2026past}
Hanmo Chen, Chenghao Xu, Xu~Yang, Xuan Chen, and Cheng Deng.
\newblock Past-and future-informed kv cache policy with salience estimation in autoregressive video diffusion.
\newblock \emph{arXiv preprint arXiv:2601.21896}, 2026{\natexlab{a}}.

\bibitem[Chen et~al.(2026{\natexlab{b}})Chen, Wang, Huang, Yang, Zhang, Xiao, Chu, Mao, Hu, Liu, et~al.]{chen2026longlive}
Yukang Chen, Luozhou Wang, Wei Huang, Shuai Yang, Bohan Zhang, Yicheng Xiao, Ruihang Chu, Weian Mao, Qixin Hu, Shaoteng Liu, et~al.
\newblock Longlive-2.0: An nvfp4 parallel infrastructure for long video generation.
\newblock \emph{arXiv preprint arXiv:2605.18739}, 2026{\natexlab{b}}.

\bibitem[Cheng et~al.(2025)Cheng, Lyu, Gu, Liu, Xu, Lu, Teng, Yang, Dong, Tang, et~al.]{cheng2025vpo}
Jiale Cheng, Ruiliang Lyu, Xiaotao Gu, Xiao Liu, Jiazheng Xu, Yida Lu, Jiayan Teng, Zhuoyi Yang, Yuxiao Dong, Jie Tang, et~al.
\newblock Vpo: Aligning text-to-video generation models with prompt optimization.
\newblock In \emph{2025 IEEE/CVF International Conference on Computer Vision (ICCV)}, pages 15636--15645. IEEE, 2025.

\bibitem[Cudlenco et~al.(2026)Cudlenco, Masala, and Leordeanu]{cudlenco2026authoring}
Nicolae Cudlenco, Mihai Masala, and Marius Leordeanu.
\newblock Authoring for living worlds: Tool-constrained llm agents for executable multi-actor scenarios, 2026.

\bibitem[Deng et~al.(2025)Deng, Pan, Diao, Luo, Cui, Lu, Shan, Qi, and Wang]{deng2025autoregressive}
Haoge Deng, Ting Pan, Haiwen Diao, Zhengxiong Luo, Yufeng Cui, Huchuan Lu, Shiguang Shan, Yonggang Qi, and Xinlong Wang.
\newblock Autoregressive video generation without vector quantization.
\newblock In \emph{International Conference on Learning Representations}, volume 2025, pages 44730--44745, 2025.

\bibitem[Feng et~al.(2025{\natexlab{a}})Feng, Lv, Cao, Xie, and Zhou]{feng2026ada}
Yuan Feng, Junlin Lv, Yukun Cao, Xike Xie, and S~Kevin Zhou.
\newblock Ada-kv: Optimizing kv cache eviction by adaptive budget allocation for efficient llm inference.
\newblock \emph{Advances in Neural Information Processing Systems}, 38:\penalty0 113152--113188, 2025{\natexlab{a}}.

\bibitem[Feng et~al.(2025{\natexlab{b}})Feng, Lv, Guo, Cao, Zhou, and Xie]{feng2025criticalkv}
Yuan Feng, Junlin Lv, Haoyu Guo, Yukun Cao, S~Kevin Zhou, and Xike Xie.
\newblock Criticalkv: Optimizing kv cache eviction from an output perturbation perspective.
\newblock \emph{arXiv preprint arXiv:2502.03805}, 2025{\natexlab{b}}.

\bibitem[Feng et~al.(2026)Feng, Wang, Xu, Qian, Wang, Hou, Liu, Sun, Liu, and Wang]{feng2026newton}
Yuxiang Feng, Juncheng Wang, Chao Xu, Yijie Qian, Huihan Wang, Wenlong Hou, Yang Liu, Baigui Sun, Yong Liu, and Shujun Wang.
\newblock Newton: Agentic planning for physically grounded video generation.
\newblock \emph{arXiv preprint arXiv:2605.18396}, 2026.

\bibitem[Fu et~al.(2026)Fu, Li, Hu, Wang, So, Dai, Chu, and Han]{fu2026ispa}
Xiaomeng Fu, Jia Li, Yiming Hu, Yong Wang, Hayden Kwok-Hay So, Jiao Dai, Xiangxiang Chu, and Jizhong Han.
\newblock Towards memory-efficient autoregressive video generation via instance-specific parametric absorption.
\newblock \emph{arXiv preprint arXiv:2607.00712}, 2026.

\bibitem[Gao et~al.(2024)Gao, Shi, Zhang, Wang, and Xiao]{gao2024vid}
Kaifeng Gao, Jiaxin Shi, Hanwang Zhang, Chunping Wang, and Jun Xiao.
\newblock Vid-gpt: Introducing gpt-style autoregressive generation in video diffusion models.
\newblock \emph{arXiv preprint arXiv:2406.10981}, 2024.

\bibitem[Gao et~al.(2025)Gao, Shi, Zhang, Wang, Xiao, and Chen]{gao2025ca2}
Kaifeng Gao, Jiaxin Shi, Hanwang Zhang, Chunping Wang, Jun Xiao, and Long Chen.
\newblock Ca2-{VDM}: Efficient autoregressive video diffusion model with causal generation and cache sharing.
\newblock In Aarti Singh, Maryam Fazel, Daniel Hsu, Simon Lacoste-Julien, Felix Berkenkamp, Tegan Maharaj, Kiri Wagstaff, and Jerry Zhu, editors, \emph{Proceedings of the 42nd International Conference on Machine Learning}, volume 267 of \emph{Proceedings of Machine Learning Research}, pages 18550--18565. PMLR, 13--19 Jul 2025.
\newblock URL \url{https://proceedings.mlr.press/v267/gao25m.html}.

\bibitem[Gu et~al.(2025)Gu, Liang, Zhao, and Diao]{gu2025obcache}
Yuzhe Gu, Xiyu Liang, Jiaojiao Zhao, and Enmao Diao.
\newblock Obcache: Optimal brain kv cache pruning for efficient long-context llm inference.
\newblock \emph{arXiv preprint arXiv:2510.07651}, 2025.

\bibitem[Guo et~al.(2026)Guo, Jia, Li, Li, Cai, Wang, Li, and Lu]{guo2026efficient}
Hang Guo, Zhaoyang Jia, Jiahao Li, Bin Li, Yuanhao Cai, Jiangshan Wang, Yawei Li, and Yan Lu.
\newblock Efficient autoregressive video diffusion with dummy head.
\newblock \emph{arXiv preprint arXiv:2601.20499}, 2026.

\bibitem[Guo et~al.(2025)Guo, Yang, He, Zhao, Wei, Yang, Huang, and Lin]{guo2025end}
Yuwei Guo, Ceyuan Yang, Hao He, Yang Zhao, Meng Wei, Zhenheng Yang, Weilin Huang, and Dahua Lin.
\newblock End-to-end training for autoregressive video diffusion via self-resampling.
\newblock \emph{arXiv preprint arXiv:2512.15702}, 2025.

\bibitem[Hao et~al.(2026)Hao, Huang, Wang, Zhang, and Yu]{hao2026deltakv}
Jitai Hao, Qiang Huang, Yaowei Wang, Min Zhang, and Jun Yu.
\newblock Deltakv: Residual-based kv cache compression via long-range similarity.
\newblock \emph{arXiv preprint arXiv:2602.08005}, 2026.

\bibitem[He et~al.(2024)He, Jiang, Zhang, Ku, Soni, Siu, Chen, Chandra, Jiang, Arulraj, et~al.]{he2024videoscore}
Xuan He, Dongfu Jiang, Ge~Zhang, Max Ku, Achint Soni, Sherman Siu, Haonan Chen, Abhranil Chandra, Ziyan Jiang, Aaran Arulraj, et~al.
\newblock Videoscore: Building automatic metrics to simulate fine-grained human feedback for video generation.
\newblock In \emph{Proceedings of the 2024 Conference on Empirical Methods in Natural Language Processing}, pages 2105--2123, 2024.

\bibitem[Henschel et~al.(2025)Henschel, Khachatryan, Poghosyan, Hayrapetyan, Tadevosyan, Wang, Navasardyan, and Shi]{henschel2025streamingt2v}
Roberto Henschel, Levon Khachatryan, Hayk Poghosyan, Daniil Hayrapetyan, Vahram Tadevosyan, Zhangyang Wang, Shant Navasardyan, and Humphrey Shi.
\newblock Streamingt2v: Consistent, dynamic, and extendable long video generation from text.
\newblock In \emph{2025 IEEE/CVF Conference on Computer Vision and Pattern Recognition (CVPR)}, pages 2568--2577. IEEE, 2025.

\bibitem[Hu et~al.(2024)Hu, Jiang, Chen, Han, Liao, Chang, and Liang]{hu2024storyagent}
Panwen Hu, Jin Jiang, Jianqi Chen, Mingfei Han, Shengcai Liao, Xiaojun Chang, and Xiaodan Liang.
\newblock Storyagent: Customized storytelling video generation via multi-agent collaboration.
\newblock \emph{arXiv preprint arXiv:2411.04925}, 2024.

\bibitem[Huang et~al.(2026{\natexlab{a}})Huang, Huang, Ning, Lin, Wang, and Liu]{huang2026genmac}
Kaiyi Huang, Yukun Huang, Xuefei Ning, Zinan Lin, Yu~Wang, and Xihui Liu.
\newblock Genmac: compositional text-to-video generation with multi-agent collaboration.
\newblock In \emph{Proceedings of the AAAI Conference on Artificial Intelligence}, volume~40, pages 5049--5057, 2026{\natexlab{a}}.

\bibitem[Huang et~al.(2026{\natexlab{b}})Huang, He, Zhou, Nie, Xia, and Huang]{huang2026vimax}
Lingxuan Huang, Sizhe He, Hengji Zhou, Liqiang Nie, Lianghao Xia, and Chao Huang.
\newblock Vimax: Agentic video generation.
\newblock \emph{arXiv preprint arXiv:2606.07649}, 2026{\natexlab{b}}.

\bibitem[Huang et~al.(2025{\natexlab{a}})Huang, Zhang, Zheng, Chao, and Ji]{huang2025layerwise}
Weizhong Huang, Yuxin Zhang, Xiawu Zheng, Fei Chao, and Rongrong Ji.
\newblock Determining layer-wise sparsity for large language models through a theoretical perspective.
\newblock In \emph{Proceedings of the 42nd International Conference on Machine Learning}, volume 267 of \emph{Proceedings of Machine Learning Research}, pages 26175--26201. PMLR, 2025{\natexlab{a}}.
\newblock URL \url{https://proceedings.mlr.press/v267/huang25ax.html}.

\bibitem[Huang et~al.(2025{\natexlab{b}})Huang, Zhang, Zheng, Chao, Ji, and Cao]{huang2025shapleymoe}
Weizhong Huang, Yuxin Zhang, Xiawu Zheng, Fei Chao, Rongrong Ji, and Liujuan Cao.
\newblock Discovering important experts for mixture-of-experts models pruning through a theoretical perspective.
\newblock In \emph{Advances in Neural Information Processing Systems}, volume~38, 2025{\natexlab{b}}.
\newblock \doi{10.52202/085713-4538}.

\bibitem[Huang et~al.(2025{\natexlab{c}})Huang, Zhang, Zheng, Liu, Lin, Yao, and Ji]{huang2025losa}
Weizhong Huang, Yuxin Zhang, Xiawu Zheng, Yang Liu, Jing Lin, Yiwu Yao, and Rongrong Ji.
\newblock Dynamic low-rank sparse adaptation for large language models.
\newblock In \emph{International Conference on Learning Representations}, 2025{\natexlab{c}}.
\newblock URL \url{https://arxiv.org/abs/2502.14816}.

\bibitem[Huang et~al.(2026{\natexlab{c}})Huang, Zhang, and Zheng]{huang2026commitkv}
Weizhong Huang, Jinchao Zhang, and Xiawu Zheng.
\newblock Commitkv: Lifecycle-aware kv cache compression via commit transitions for multi-turn agents.
\newblock \emph{arXiv preprint arXiv:2608.07855}, 2026{\natexlab{c}}.

\bibitem[Huang et~al.(2025{\natexlab{d}})Huang, Li, He, Zhou, and Shechtman]{huang2026self}
Xun Huang, Zhengqi Li, Guande He, Mingyuan Zhou, and Eli Shechtman.
\newblock Self forcing: Bridging the train-test gap in autoregressive video diffusion.
\newblock \emph{Advances in Neural Information Processing Systems}, 38:\penalty0 167283--167308, 2025{\natexlab{d}}.

\bibitem[Huang et~al.(2025{\natexlab{e}})Huang, Wang, Lin, Kim, Omidshafiei, Yoon, Zhang, and Bansal]{huang2025planning}
Yidong Huang, Zun Wang, Han Lin, Dong-Ki Kim, Shayegan Omidshafiei, Jaehong Yoon, Yue Zhang, and Mohit Bansal.
\newblock Planning with sketch-guided verification for physics-aware video generation.
\newblock \emph{arXiv preprint arXiv:2511.17450}, 2025{\natexlab{e}}.

\bibitem[Huang et~al.(2024)Huang, He, Yu, Zhang, Si, Jiang, Zhang, Wu, Jin, Chanpaisit, et~al.]{huang2024vbench}
Ziqi Huang, Yinan He, Jiashuo Yu, Fan Zhang, Chenyang Si, Yuming Jiang, Yuanhan Zhang, Tianxing Wu, Qingyang Jin, Nattapol Chanpaisit, et~al.
\newblock Vbench: Comprehensive benchmark suite for video generative models.
\newblock In \emph{2024 IEEE/CVF Conference on Computer Vision and Pattern Recognition (CVPR)}, pages 21807--21818. IEEE, 2024.

\bibitem[Huang et~al.(2025{\natexlab{f}})Huang, Zhang, Xu, He, Yu, Dong, Ma, Chanpaisit, Si, Jiang, et~al.]{huang2025vbench++}
Ziqi Huang, Fan Zhang, Xiaojie Xu, Yinan He, Jiashuo Yu, Ziyue Dong, Qianli Ma, Nattapol Chanpaisit, Chenyang Si, Yuming Jiang, et~al.
\newblock Vbench++: Comprehensive and versatile benchmark suite for video generative models.
\newblock \emph{IEEE Transactions on Pattern Analysis and Machine Intelligence}, 2025{\natexlab{f}}.

\bibitem[Ji et~al.(2026)Ji, Zhong, Zhang, Yang, Jin, Qin, Luo, Mao, Liu, and Li]{ji2026forcing}
Yicheng Ji, Zhizhou Zhong, Jun Zhang, Qin Yang, XiTai Jin, Ying Qin, Wenhan Luo, Shuiyang Mao, Wei Liu, and Huan Li.
\newblock Forcing-kv: Hybrid kv cache compression for efficient autoregressive video diffusion models.
\newblock \emph{arXiv preprint arXiv:2605.09681}, 2026.

\bibitem[Li et~al.(2026{\natexlab{a}})Li, Shah, and Shang]{li2026packcache}
Kunyang Li, Mubarak Shah, and Yuzhang Shang.
\newblock Packcache: A training-free acceleration method for unified autoregressive video generation via compact kv-cache.
\newblock \emph{arXiv preprint arXiv:2601.04359}, 2026{\natexlab{a}}.

\bibitem[Li et~al.(2026{\natexlab{b}})Li, Pan, Luan, Gao, and Alahi]{li2026stable}
Wuyang Li, Wentao Pan, Po-Chien Luan, Yang Gao, and Alexandre Alahi.
\newblock Stable video infinity: Infinite-length video generation with error recycling.
\newblock In \emph{International Conference on Learning Representations}, volume 2026, pages 23406--23432, 2026{\natexlab{b}}.

\bibitem[Li et~al.(2026{\natexlab{c}})Li, Li, and Lan]{li2026momentkv}
Yu~Li, Binxu Li, and Tian Lan.
\newblock Momentkv: Closing the directional gap in kv cache eviction for long-context inference.
\newblock \emph{arXiv preprint arXiv:2606.01563}, 2026{\natexlab{c}}.

\bibitem[Liang et~al.(2025)Liang, Zhang, Zhou, Huang, Li, Zhang, Wu, Wang, Luo, Liao, et~al.]{liang2025univa}
Zhengyang Liang, Daoan Zhang, Huichi Zhou, Rui Huang, Bobo Li, Yuechen Zhang, Shengqiong Wu, Xiaohan Wang, Jiebo Luo, Lizi Liao, et~al.
\newblock Univa: Universal video agent towards open-source next-generation video generalist.
\newblock \emph{arXiv preprint arXiv:2511.08521}, 2025.

\bibitem[Lin et~al.(2024)Lin, Zala, Cho, and Bansal]{lin2024videodirectorgpt}
Han Lin, Abhay Zala, Jaemin Cho, and Mohit Bansal.
\newblock Videodirectorgpt: Consistent multi-scene video generation via llm-guided planning.
\newblock In \emph{First Conference on Language Modeling}, 2024.
\newblock URL \url{https://openreview.net/forum?id=sKNIjS2brr}.

\bibitem[Lin et~al.(2025)Lin, Yang, He, Jiang, Ren, Xia, Zhao, Xiao, and Jiang]{lin2026autoregressive}
Shanchuan Lin, Ceyuan Yang, Hao He, Jianwen Jiang, Yuxi Ren, Xin Xia, Yang Zhao, Xuefeng Xiao, and Lu~Jiang.
\newblock Autoregressive adversarial post-training for real-time interactive video generation.
\newblock \emph{Advances in Neural Information Processing Systems}, 38:\penalty0 41061--41086, 2025.

\bibitem[Liu et~al.(2025)Liu, Wang, Cai, Zhang, Zhan, and Duan]{liu2025video}
Fangfu Liu, Hanyang Wang, Yimo Cai, Kaiyan Zhang, Xiaohang Zhan, and Yueqi Duan.
\newblock Video-t1: Test-time scaling for video generation.
\newblock In \emph{2025 IEEE/CVF International Conference on Computer Vision (ICCV)}, pages 18671--18681. IEEE, 2025.

\bibitem[Liu et~al.(2026)Liu, Hu, Xu, Shan, and Lu]{liu2026rolling}
Kunhao Liu, Wenbo Hu, Jiale Xu, Ying Shan, and Shijian Lu.
\newblock Rolling forcing: Autoregressive long video diffusion in real time.
\newblock In \emph{International Conference on Learning Representations}, volume 2026, pages 91177--91196, 2026.

\bibitem[Long et~al.(2026)Long, Wan, Nakhost, Lee, Pfister, and Arik]{long2026vista}
Do~Xuan Long, Xingchen Wan, Hootan Nakhost, Chen-Yu Lee, Tomas Pfister, and Sercan~{\"O} Arik.
\newblock Vista: A test-time self-improving video generation agent.
\newblock In \emph{Proceedings of the IEEE/CVF conference on computer vision and pattern recognition}, pages 6021--6032, 2026.

\bibitem[Luo et~al.(2026{\natexlab{a}})Luo, Liu, Wang, Liu, Chen, Wang, Zhu, Gao, Hu, Sun, et~al.]{luo2026future}
Jiayi Luo, Qiyan Liu, Tengyang Wang, JunHao Liu, Jiayu Chen, Cong Wang, Hanxin Zhu, Chen Gao, Xiaobin Hu, Qingyun Sun, et~al.
\newblock Future forcing: Future-aware training-free kv cache policy for autoregressive video generation.
\newblock \emph{arXiv preprint arXiv:2605.30083}, 2026{\natexlab{a}}.

\bibitem[Luo et~al.(2026{\natexlab{b}})Luo, Chen, Huang, Yin, Lin, Huang, Fu, Ji, Zheng, and Luo]{luo2026quota}
Yongdong Luo, Wang Chen, Weizhong Huang, Shukang Yin, Haojia Lin, Jinfa Huang, Chaoyou Fu, Jiayi Ji, Xiawu Zheng, and Jiebo Luo.
\newblock {QuoTA}: Query-oriented token assignment via {CoT} query decouple for long video comprehension.
\newblock \emph{Proceedings of the AAAI Conference on Artificial Intelligence}, 40\penalty0 (29):\penalty0 24160--24168, 2026{\natexlab{b}}.
\newblock \doi{10.1609/aaai.v40i29.39595}.
\newblock URL \url{https://ojs.aaai.org/index.php/AAAI/article/view/39595}.

\bibitem[Lv et~al.(2026)Lv, Shi, Huang, Gong, Ren, and Wang]{lv2026light}
Chengtao Lv, Yumeng Shi, Yushi Huang, Ruihao Gong, Shen Ren, and Wenya Wang.
\newblock Light forcing: Accelerating autoregressive video diffusion via sparse attention.
\newblock In \emph{Forty-third International Conference on Machine Learning}, 2026.
\newblock URL \url{https://openreview.net/forum?id=NixeX51WQA}.

\bibitem[Ma et~al.(2026)Ma, Zheng, Xu, Xu, Ling, Zheng, Kuang, Li, WANG, Xiao, Chao, and Ji]{ma2026flow}
Yuexiao Ma, Xuzhe Zheng, Jing Xu, Xiwei Xu, Feng Ling, Xiawu Zheng, Huafeng Kuang, Huixia Li, XING WANG, Xuefeng Xiao, Fei Chao, and Rongrong Ji.
\newblock Flow caching for autoregressive video generation.
\newblock In \emph{The Fourteenth International Conference on Learning Representations}, 2026.
\newblock URL \url{https://openreview.net/forum?id=vko4DuhKbh}.

\bibitem[Mao et~al.(2026)Mao, Rui, Ying, Zheng, Li, Chi, and Zhang]{mao2026packforcing}
Xiaofeng Mao, Shaohao Rui, Kaining Ying, Bo~Zheng, Chuanhao Li, Mingmin Chi, and Kaipeng Zhang.
\newblock Packforcing: Short video training suffices for long video sampling and long context inference.
\newblock \emph{arXiv preprint arXiv:2603.25730}, 2026.

\bibitem[Meng et~al.(2026)Meng, Luo, Li, Jiang, Gao, Chen, Li, and Zhang]{meng2026tethercache}
Yu~Meng, Xiangyang Luo, Letian Li, Wenyuan Jiang, Chen Gao, Xinlei Chen, Yong Li, and Xiao-Ping Zhang.
\newblock Tethercache: Stabilizing autoregressive long-form video generation with gated recall and trusted alignment.
\newblock \emph{arXiv preprint arXiv:2606.13035}, 2026.

\bibitem[Polyak et~al.(2024)Polyak, Zohar, Brown, Tjandra, Sinha, Lee, Vyas, Shi, Ma, Chuang, et~al.]{polyak2024movie}
Adam Polyak, Amit Zohar, Andrew Brown, Andros Tjandra, Animesh Sinha, Ann Lee, Apoorv Vyas, Bowen Shi, Chih-Yao Ma, Ching-Yao Chuang, et~al.
\newblock Movie gen: A cast of media foundation models.
\newblock \emph{arXiv preprint arXiv:2410.13720}, 2024.

\bibitem[Radford et~al.(2021)Radford, Kim, Hallacy, Ramesh, Goh, Agarwal, Sastry, Askell, Mishkin, Clark, Krueger, and Sutskever]{radford2021clip}
Alec Radford, Jong~Wook Kim, Chris Hallacy, Aditya Ramesh, Gabriel Goh, Sandhini Agarwal, Girish Sastry, Amanda Askell, Pamela Mishkin, Jack Clark, Gretchen Krueger, and Ilya Sutskever.
\newblock Learning transferable visual models from natural language supervision.
\newblock In Marina Meila and Tong Zhang, editors, \emph{Proceedings of the 38th International Conference on Machine Learning}, volume 139 of \emph{Proceedings of Machine Learning Research}, pages 8748--8763. PMLR, 18--24 Jul 2021.
\newblock URL \url{https://proceedings.mlr.press/v139/radford21a.html}.

\bibitem[Samuel et~al.(2026)Samuel, Tzachor, Levy, Green, Chechik, and Ben-Ari]{samuel2026fastar}
Dvir Samuel, Issar Tzachor, Matan Levy, Michael Green, Gal Chechik, and Rami Ben-Ari.
\newblock Fast-ar: Fast autoregressive video diffusion and world models with temporal cache compression and sparse attention.
\newblock In \emph{Forty-third International Conference on Machine Learning}, 2026.
\newblock URL \url{https://openreview.net/forum?id=UxgMJP0JKY}.

\bibitem[Soni et~al.(2024)Soni, Venkataraman, Chandra, Fischmeister, Liang, Dai, and Yang]{soni2024videoagent}
Achint Soni, Sreyas Venkataraman, Abhranil Chandra, Sebastian Fischmeister, Percy Liang, Bo~Dai, and Sherry Yang.
\newblock Videoagent: Self-improving video generation.
\newblock \emph{arXiv preprint arXiv:2410.10076}, 2024.

\bibitem[Teed and Deng(2020)]{teed2020raft}
Zachary Teed and Jia Deng.
\newblock Raft: Recurrent all-pairs field transforms for optical flow.
\newblock In \emph{European conference on computer vision}, pages 402--419. Springer, 2020.

\bibitem[Wan et~al.(2025)Wan, Wang, Ai, Wen, Mao, Xie, Chen, Yu, Zhao, Yang, et~al.]{wan2025wan}
Team Wan, Ang Wang, Baole Ai, Bin Wen, Chaojie Mao, Chen-Wei Xie, Di~Chen, Feiwu Yu, Haiming Zhao, Jianxiao Yang, et~al.
\newblock Wan: Open and advanced large-scale video generative models.
\newblock \emph{arXiv preprint arXiv:2503.20314}, 2025.

\bibitem[Wang et~al.(2026)Wang, Zhao, Guo, Wang, Guo, Zhu, Li, and Yue]{wang2026precisecache}
Jiangshan Wang, Kang Zhao, Jiayi Guo, Jiayu Wang, Hang Guo, Chenyang Zhu, Xiu Li, and Xiangyu Yue.
\newblock Precisecache: Precise feature caching for efficient and high-fidelity video generation.
\newblock \emph{arXiv preprint arXiv:2603.00976}, 2026.

\bibitem[Wang et~al.(2025)Wang, Zhang, Li, Tan, Pang, Du, Sun, and Yang]{wang2025error}
Jing Wang, Fengzhuo Zhang, Xiaoli Li, Vincent~YF Tan, Tianyu Pang, Chao Du, Aixin Sun, and Zhuoran Yang.
\newblock Error analyses of auto-regressive video diffusion models: A unified framework.
\newblock \emph{arXiv preprint arXiv:2503.10704}, 2025.

\bibitem[Wang et~al.(2024{\natexlab{a}})Wang, Du, Zhao, Yuan, Wang, Liang, Zhao, Lu, Li, Gao, et~al.]{wang2024aesopagent}
Jiuniu Wang, Zehua Du, Yuyuan Zhao, Bo~Yuan, Kexiang Wang, Jian Liang, Yaxi Zhao, Yihen Lu, Gengliang Li, Junlong Gao, et~al.
\newblock Aesopagent: Agent-driven evolutionary system on story-to-video production.
\newblock \emph{arXiv preprint arXiv:2403.07952}, 2024{\natexlab{a}}.

\bibitem[Wang et~al.(2024{\natexlab{b}})Wang, Xiong, Zhou, Lin, Zhao, Kang, Feng, and Liu]{wang2024loong}
Yuqing Wang, Tianwei Xiong, Daquan Zhou, Zhijie Lin, Yang Zhao, Bingyi Kang, Jiashi Feng, and Xihui Liu.
\newblock Loong: Generating minute-level long videos with autoregressive language models.
\newblock \emph{arXiv preprint arXiv:2410.02757}, 2024{\natexlab{b}}.

\bibitem[Wu et~al.(2025)Wu, Zhu, and Shou]{wu2025automated}
Weijia Wu, Zeyu Zhu, and Mike~Zheng Shou.
\newblock Automated movie generation via multi-agent cot planning.
\newblock \emph{arXiv preprint arXiv:2503.07314}, 2025.

\bibitem[Xi et~al.(2026)Xi, Yang, Zhao, Li, Cai, Li, Lin, Zhang, Zhang, Li, Xu, Wu, Xu, Stoica, Han, and Keutzer]{xi2026quant}
Haocheng Xi, Shuo Yang, Yilong Zhao, Muyang Li, Han Cai, Xingyang Li, Yujun Lin, Zhuoyang Zhang, Jintao Zhang, Xiuyu Li, Zhiying Xu, Jun Wu, Chenfeng Xu, Ion Stoica, Song Han, and Kurt Keutzer.
\newblock Quant videogen: Auto-regressive long video generation via 2-bit {KV}-cache quantization.
\newblock In \emph{Forty-third International Conference on Machine Learning}, 2026.
\newblock URL \url{https://openreview.net/forum?id=r9UJgVH1Fx}.

\bibitem[Xie et~al.(2025)Xie, Xu, Hong, Tan, Liu, Liu, Kaufman, and Zhou]{xie2025progressive}
Desai Xie, Zhan Xu, Yicong Hong, Hao Tan, Difan Liu, Feng Liu, Arie Kaufman, and Yang Zhou.
\newblock Progressive autoregressive video diffusion models.
\newblock In \emph{2025 IEEE/CVF Conference on Computer Vision and Pattern Recognition Workshops (CVPRW)}, pages 6312--6322. IEEE, 2025.

\bibitem[Xie et~al.(2024)Xie, Tang, Tan, Klein, Bissyand, and Ezzini]{xie2024dreamfactory}
Zhifei Xie, Daniel Tang, Dingwei Tan, Jacques Klein, Tegawend~F Bissyand, and Saad Ezzini.
\newblock Dreamfactory: Pioneering multi-scene long video generation with a multi-agent framework.
\newblock \emph{arXiv preprint arXiv:2408.11788}, 2024.

\bibitem[Xing et~al.(2024)Xing, Fox, Zeng, Pan, Elgharib, Theobalt, and Chen]{xing2024live2diff}
Zhening Xing, Gereon Fox, Yanhong Zeng, Xingang Pan, Mohamed Elgharib, Christian Theobalt, and Kai Chen.
\newblock Live2diff: Live stream translation via uni-directional attention in video diffusion models.
\newblock \emph{arXiv preprint arXiv:2407.08701}, 2024.

\bibitem[Xu et~al.(2026)Xu, Huang, Cheng, Yang, Xu, Wang, Duan, Yang, Jin, Li, et~al.]{xu2026visionreward}
Jiazheng Xu, Yu~Huang, Jiale Cheng, Yuanming Yang, Jiajun Xu, Yuan Wang, Wenbo Duan, Shen Yang, Qunlin Jin, Shurun Li, et~al.
\newblock Visionreward: Fine-grained multi-dimensional human preference learning for image and video generation.
\newblock In \emph{Proceedings of the AAAI Conference on Artificial Intelligence}, volume~40, pages 11269--11277, 2026.

\bibitem[Xu et~al.(2025)Xu, Wang, Wang, Li, Shi, Yang, Wang, Hu, Yu, and Zhang]{xu2025filmagent}
Zhenran Xu, Longyue Wang, Jifang Wang, Zhouyi Li, Senbao Shi, Xue Yang, Yiyu Wang, Baotian Hu, Jun Yu, and Min Zhang.
\newblock Filmagent: A multi-agent framework for end-to-end film automation in virtual 3d spaces.
\newblock \emph{arXiv preprint arXiv:2501.12909}, 2025.

\bibitem[Yang et~al.(2026{\natexlab{a}})Yang, Li, Qi, Zhou, Wu, and Liu]{yang2026scmapr}
Chengyi Yang, Pengzhen Li, Jiayin Qi, Aimin Zhou, Ji~Wu, and Ji~Liu.
\newblock Scmapr: Self-correcting multi-agent prompt refinement for complex-scenario text-to-video generation.
\newblock In \emph{Findings of the Association for Computational Linguistics: ACL 2026}, pages 6942--6973, 2026{\natexlab{a}}.

\bibitem[Yang et~al.(2026{\natexlab{b}})Yang, Tang, Zhen, Zhang, and Lv]{yang2026jacap}
Jiaming Yang, Chenwei Tang, Liangli Zhen, Chenyang Zhang, and Jiancheng Lv.
\newblock Jacap: Robust kv cache eviction via jacobian-based nonlinear information capacity preservation, 2026{\natexlab{b}}.

\bibitem[Yang et~al.(2026{\natexlab{c}})Yang, Huang, Chu, Xiao, Zhao, Wang, Li, Xie, CHEN, Lu, Han, and Chen]{yang2026longlive}
Shuai Yang, Wei Huang, Ruihang Chu, Yicheng Xiao, Yuyang Zhao, Xianbang Wang, Muyang Li, Enze Xie, YINGCONG CHEN, Yao Lu, Song Han, and Yukang Chen.
\newblock Longlive: Real-time interactive long video generation.
\newblock In \emph{International Conference on Learning Representations}, volume 2026, pages 90037--90059, 2026{\natexlab{c}}.

\bibitem[Yang et~al.(2026{\natexlab{d}})Yang, Liao, Mei, Wang, Yang, Wen, Zhang, Li, Lv, Chen, et~al.]{yang2026spiral}
Yu~Yang, Yue Liao, Jianbiao Mei, Baisen Wang, Xuemeng Yang, Licheng Wen, Jiangning Zhang, Xiangtai Li, Liang Lv, Hanlin Chen, et~al.
\newblock Spiral: Self-evolving action-conditioned video generation via reflective planning agents.
\newblock \emph{arXiv preprint arXiv:2603.08403}, 2026{\natexlab{d}}.

\bibitem[Yin et~al.(2025)Yin, Zhang, Zhang, Freeman, Durand, Shechtman, and Huang]{yin2025slow}
Tianwei Yin, Qiang Zhang, Richard Zhang, William~T Freeman, Fredo Durand, Eli Shechtman, and Xun Huang.
\newblock From slow bidirectional to fast autoregressive video diffusion models.
\newblock In \emph{2025 IEEE/CVF Conference on Computer Vision and Pattern Recognition (CVPR)}, pages 22963--22974. IEEE, 2025.

\bibitem[Yu et~al.(2025)Yu, Wu, Hu, Hu, Sun, Lyu, Wang, Ma, Ma, Wang, et~al.]{yu2025videossm}
Yifei Yu, Xiaoshan Wu, Xinting Hu, Tao Hu, Yangtian Sun, Xiaoyang Lyu, Bo~Wang, Lin Ma, Yuewen Ma, Zhongrui Wang, et~al.
\newblock Videossm: Autoregressive long video generation with hybrid state-space memory.
\newblock \emph{arXiv preprint arXiv:2512.04519}, 2025.

\bibitem[Yuan et~al.(2024)Yuan, Liu, Cao, Sun, Jia, Chen, Li, Lin, Yuan, He, et~al.]{yuan2024mora}
Zhengqing Yuan, Yixin Liu, Yihan Cao, Weixiang Sun, Haolong Jia, Ruoxi Chen, Zhaoxu Li, Bin Lin, Li~Yuan, Lifang He, et~al.
\newblock Mora: Enabling generalist video generation via a multi-agent framework.
\newblock \emph{arXiv preprint arXiv:2403.13248}, 2024.

\bibitem[Zeng et~al.(2025)Zeng, Cai, Chen, Lv, and Wang]{zeng2025coagent}
Qinglin Zeng, Kaitong Cai, Ruiqi Chen, Qinhan Lv, and Keze Wang.
\newblock Coagent: Collaborative planning and consistency agent for coherent video generation.
\newblock \emph{arXiv preprint arXiv:2512.22536}, 2025.

\bibitem[Zhang et~al.(2026{\natexlab{a}})Zhang, Cong, Zhou, Zhong, Xu, Mao, Liu, and Li]{zhang2026kvpo}
Ruicheng Zhang, Kaixi Cong, Jun Zhou, Zhizhou Zhong, Zunnan Xu, Shuiyang Mao, Wei Liu, and Xiu Li.
\newblock Kvpo: Ode-native grpo for autoregressive video alignment via kv semantic exploration.
\newblock \emph{arXiv preprint arXiv:2605.14278}, 2026{\natexlab{a}}.

\bibitem[Zhang et~al.(2026{\natexlab{b}})Zhang, Huang, Ma, Zhong, Zheng, and Ji]{zhang2026spt}
Yuxin Zhang, Weizhong Huang, Yuexiao Ma, Yunshan Zhong, Xiawu Zheng, and Rongrong Ji.
\newblock Sparsity-aware prompt tuning: A simple and effective way to fine-tune high-sparsity {LLMs}.
\newblock In \emph{Conference on Parsimony and Learning}, volume 328 of \emph{Proceedings of Machine Learning Research}, pages 644--657. PMLR, 2026{\natexlab{b}}.
\newblock URL \url{https://proceedings.mlr.press/v328/zhang26b.html}.

\bibitem[Zhao et~al.(2026)Zhao, Chi, Zhang, Ma, Li, Fang, Tang, Gao, and Wu]{zhao2026densitykv}
Wenqu Zhao, Xuemin Chi, Xin Zhang, Guoqing Ma, Baorun Li, Jianjie Fang, Peizhi Tang, Chen Gao, and Wei Wu.
\newblock Densitykv: Density-guided kv cache compression for long video generation.
\newblock \emph{arXiv preprint arXiv:2608.27922}, 2026.

\bibitem[Zheng et~al.(2023)Zheng, Chiang, Sheng, Zhuang, Wu, Zhuang, Lin, Li, Li, Xing, et~al.]{zheng2023judging}
Lianmin Zheng, Wei-Lin Chiang, Ying Sheng, Siyuan Zhuang, Zhanghao Wu, Yonghao Zhuang, Zi~Lin, Zhuohan Li, Dacheng Li, Eric Xing, et~al.
\newblock Judging llm-as-a-judge with mt-bench and chatbot arena.
\newblock \emph{Advances in neural information processing systems}, 36:\penalty0 46595--46623, 2023.

\bibitem[Zhu et~al.(2026)Zhu, Zhao, He, Su, Li, and Zhu]{zhu2026causal}
Hongzhou Zhu, Min Zhao, Guande He, Hang Su, Chongxuan Li, and Jun Zhu.
\newblock Causal forcing: Autoregressive diffusion distillation done right for high-quality real-time interactive video generation.
\newblock In \emph{Forty-third International Conference on Machine Learning}, 2026.
\newblock URL \url{https://openreview.net/forum?id=BYInOck3gr}.

\bibitem[Zhuang et~al.(2024)Zhuang, Li, Chen, Wang, Liu, Qiao, and Wang]{zhuang2024vlogger}
Shaobin Zhuang, Kunchang Li, Xinyuan Chen, Yaohui Wang, Ziwei Liu, Yu~Qiao, and Yali Wang.
\newblock Vlogger: Make your dream a vlog.
\newblock In \emph{2024 IEEE/CVF Conference on Computer Vision and Pattern Recognition (CVPR)}, pages 8806--8817. IEEE, 2024.

\bibitem[Zweiger et~al.(2026)Zweiger, Fu, Guo, and Kim]{zweiger2026attentionmatching}
Adam Zweiger, Xinghong Fu, Han Guo, and Yoon Kim.
\newblock Fast kv compaction via attention matching.
\newblock \emph{arXiv preprint arXiv:2602.16284}, 2026.

\end{thebibliography}
\bibliographystyle{plainnat}

\appendix
\startcontents
\printcontents{}{1}{\section*{Appendix}}

\section{Detailed Algorithm}
\label{app:detailed_algorithm}

Algorithm~\ref{alg:SemanticKV} summarizes the cache-update procedure of \method{}, combining constraint-conditioned query selection, groupwise attention scoring, and budgeted top-$k$ retention. Attention weights are averaged within each semantic group and maximized across groups to rank KV entries. The procedure preserves backbone-required states and skips compression when all available entries fit within the budget. Otherwise, selected keys, values, and positions are packed in their original logical order without modification.

\begin{algorithm}[htbp]
\caption{\method{}: core steps of a semantically aligned cache update}
\label{alg:SemanticKV}
\small
\begin{algorithmic}[1]
\Require Available cache $(\mathbf{K}_t,\mathbf{V}_t)$ indexed by $\mathcal{H}_t$; semantic constraints $\mathcal{R}$; candidate prompt $\mathbf{q}^{i}$
\Require Observed self-attention queries and text cross-attention; unit budget $B_t$ determined by compression ratio $r$; query limit $n_q$
\Ensure Compressed cache $(\widetilde{\mathbf{K}}_t,\widetilde{\mathbf{V}}_t)$ within $B_t$
\State Identify mandatory entries $\mathcal{A}_t$ and check budget feasibility
\If{all available entries fit within $B_t$}
    \State \Return The unchanged cache $(\mathbf{K}_t,\mathbf{V}_t)$
\EndIf
\State Match each constraint $r_a$ to prompt-token positions $\mathcal{T}_a$
\State Compute routing scores $g_a(x)$ using Eq.~\ref{eq:atom_routing}
\State Collect self-attention queries at up to $n_q$ top-ranked positions per reliably matched constraint; form index groups $\mathcal{P}_a$
\State Let $\mathcal{I}$ index all nonempty query groups used for scoring
\State Compute masked cache-attention weights $\alpha_{pj}$ using Eq.~\ref{eq:semantic_attention}
\State Compute group means $w_{aj}$ and semantic importance $s_j$ using Eq.~\ref{eq:semantic_score}
\State Determine $k_t$ from $B_t$, accounting for mandatory states, positions, metadata, and allocation granularity
\State Select retained indices $\mathcal{S}_t$ using Eq.~\ref{eq:semantic_retention}
\State \Return $(\mathbf{K}_{\mathcal{S}_t},\mathbf{V}_{\mathcal{S}_t})$ packed in original logical order with unchanged positions
\end{algorithmic}
\end{algorithm}

\section{Details of Models, Baselines, Metrics and Benchmarks}
\subsection{Details of Models}
\label{sec:Details_of_Models}
\paragraph{Basic information.}
We evaluate four autoregressive video diffusion models spanning different model scales and generation designs.
\begin{itemize}
\item \textbf{LongLive}~\citep{yang2026longlive}.
LongLive is a 1.3B-parameter autoregressive video diffusion model for real-time long-video generation. It adopts frame-level causal generation with KV caching and introduces a frame sink, short attention window, and KV recaching to support efficient long-horizon and interactive generation.

\item \textbf{LongLive 2.0}~\citep{chen2026longlive}.
LongLive 2.0 is a 5B-parameter autoregressive video diffusion model designed for scalable long-video generation. It extends the LongLive framework to a larger backbone and supports long-horizon generation with KV-cached causal attention and efficient sequence-parallel inference.

\item \textbf{Self Forcing}~\citep{huang2026self}.
Self Forcing is a 1.3B-parameter autoregressive video diffusion model based on Wan2.1-T2V-1.3B~\citep{wan2025wan}. It performs causal video generation with temporal KV caching and supports both frame-wise and chunk-wise autoregressive generation, making it a representative backbone for efficient short-video generation.

\item \textbf{Causal Forcing}~\citep{zhu2026causal}.
Causal Forcing is a 1.3B-parameter autoregressive video diffusion model derived from Wan2.1-T2V-1.3B~\citep{wan2025wan}. It performs few-step causal generation with temporal KV caching and provides an additional autoregressive formulation for evaluating the generalization of our method.
\end{itemize}

\paragraph{Model-specific inference configurations.}
We use the following inference configurations for the four backbones. Resolutions are reported as width $\times$ height, and chunk sizes are measured in latent frames. The video frame rates below describe playback rather than the generation throughput reported in our efficiency evaluation.
\begin{itemize}
\item \textbf{LongLive (v1).} We use BF16 precision, $832\times480$ resolution, a video frame rate of 16 FPS, four denoising steps, and three latent frames per autoregressive chunk.

\item \textbf{LongLive 2.0.} We use BF16 precision, $1280\times704$ resolution, a video frame rate of 24 FPS, four denoising steps, and eight latent frames per autoregressive chunk.

\item \textbf{Self Forcing.} We use BF16 precision, $832\times480$ resolution, a video frame rate of 16 FPS, four denoising steps, and three latent frames per autoregressive chunk.

\item \textbf{Causal Forcing.} We use BF16 precision, $832\times480$ resolution, a video frame rate of 16 FPS, four denoising steps, and three latent frames per autoregressive chunk.
\end{itemize}

\subsection{Details of Baselines}
\label{sec:Details_of_Baselines}
\begin{itemize}
\item \textbf{Full KV.} Full KV serves as the uncompressed reference and applies no additional cache eviction. Its configuration should specify the backbone's native attention mask, context window, sink states, and re-caching behavior, so that the reference cache footprint is unambiguous.

\item \textbf{R-KV~\citep{rkv}.} R-KV selects cached tokens using importance and redundancy. Its adaptation to video generation must specify the source of the importance scores, the treatment of video-token positions, and the cache-update schedule. Any changes required by chunk-wise denoising should be documented separately from the original selection rule.

\item \textbf{Dummy Forcing~\citep{guo2026efficient}.} Dummy Forcing allocates historical context according to attention-head locality. Reproduction requires the head-classification procedure, profiling inputs, retained context for each head category, and the frequency of classification. Profiling and cache-layout conversion costs belong in the measured inference time.

\item \textbf{Forcing-KV~\citep{ji2026forcing}.} Forcing-KV applies different retention policies to static and dynamic heads. The configuration should disclose the head-partition criterion, category-specific retention settings, protected context, and compression schedule. The realized cache footprint should be reported in addition to the method's nominal compression parameters.
\end{itemize}

\subsection{Details of Benchmarks}
\label{sec:Details_of_Benchmarks}
We evaluate video generation quality using benchmarks covering short-video quality, long-horizon consistency, and human preference.
\begin{itemize}
\item \textbf{VBench}~\citep{huang2024vbench}.
VBench is a comprehensive benchmark for video generation that evaluates 16 fine-grained dimensions covering visual quality, temporal consistency, motion, and semantic alignment. We use all 946 official prompts for 5-second video evaluation.

\item \textbf{VBenchLong}~\citep{huang2025vbench++}.
VBenchLong extends fine-grained video evaluation to long-duration generation, emphasizing temporal consistency, motion quality, and visual stability over extended sequences. We evaluate 30-second and 60-second videos using 128 MovieGen prompts with five random seeds per prompt.

\item \textbf{VisionReward}~\citep{xu2026visionreward}.
VisionReward is a fine-grained human-preference reward model for image and video generation. It evaluates multiple aspects of perceptual quality and instruction alignment and aggregates them into an overall preference score. We follow the Causal Forcing evaluation protocol and use its 100-prompt evaluation set.
\end{itemize}

\subsection{Details of Evaluation Metrics}
\label{app:video_metrics}

We evaluate both generation efficiency and video quality. The efficiency metrics include frames per second (FPS) and inference speedup, while the quality evaluation covers chunk continuity, temporal dynamics, visual quality, temporal consistency, and human-preference-aligned generation quality.

\paragraph{Frames per Second and Speedup.}
We measure generation throughput using frames per second (FPS). Following prior work, the runtime is measured within the diffusion transformer (DiT), excluding the VAE-related encoding and decoding overhead. Given a generated video containing $F$ frames and a DiT inference time of $T$, FPS is computed as
\begin{equation}
\mathrm{FPS} = \frac{F}{T}.
\end{equation}
We further report the inference speedup relative to the corresponding Full KV configuration:
\begin{equation}
\mathrm{Speedup}
=
\frac{\mathrm{FPS}_{\mathrm{method}}}
{\mathrm{FPS}_{\mathrm{FullKV}}}.
\end{equation}
Therefore, Full KV has a speedup of $1.00\times$ by definition.

\paragraph{Chunk Discontinuity.}
Autoregressive video diffusion models generate videos chunk by chunk, making visual continuity at chunk boundaries particularly important. We quantify abrupt transitions between consecutive chunks using the chunk discontinuity metric. Given a video with $F$ frames generated in $K$ autoregressive chunks, we first compute the optical-flow difference between every pair of adjacent frames using RAFT~\citep{teed2020raft}:
\begin{equation}
\delta_t = \Delta_{\mathrm{RAFT}}(I_t,I_{t+1}),
\qquad t=1,\ldots,F-1,
\end{equation}
where $I_t$ denotes the $t$-th frame. We then identify the $K-1$ largest optical-flow differences, corresponding to the strongest temporal discontinuities, and normalize their average by the mean optical-flow difference over the entire video:
\begin{equation}
\mathrm{ChunkDisc.}
=
\frac{
\frac{1}{K-1}
\sum_{\delta \in \mathrm{Top}_{K-1}(\{\delta_t\})}\delta
}{
\frac{1}{F-1}\sum_{t=1}^{F-1}\delta_t
}.
\end{equation}
A lower score indicates smoother transitions across autoregressive chunks and fewer abrupt visual changes at chunk boundaries.

\paragraph{Dynamic Degree.}
Dynamic Degree measures the amount of motion and temporal variation in the generated video. We obtain this score using the official VBench/VBenchLong evaluation pipeline. A higher Dynamic Degree indicates richer and more pronounced motion, whereas a lower value generally corresponds to more static video content. This metric is particularly relevant for KV-cache compression, since excessively removing historical context may preserve individual-frame quality while suppressing temporal dynamics.

\paragraph{VBench Metrics.}
For short-video generation, we evaluate 5-second videos using the official VBench~\citep{huang2024vbench} evaluation pipeline. VBench evaluates generated videos along 16 dimensions that characterize different aspects of visual and temporal quality. Table~\ref{tab:main_result_short} reports the total score and representative visual and temporal metrics, complemented by the main-paper figures. Table~\ref{tab:main_result_backbone} in Appendix~\ref{app:full_benchmark_results} reports all 16 VBench dimensions, while Table~\ref{tab:main_result_short_full} reports the six dimensions shown in Figure~\ref{fig:cache_setting_radar}.

\textit{Imaging Quality} evaluates the visual fidelity and perceptual quality of individual frames.
\textit{Subject Consistency} measures whether the appearance and identity of the primary subject remain stable throughout the video.
\textit{Background Consistency} evaluates the temporal stability of the background scene.
\textit{Motion Smoothness} measures whether motion evolves smoothly over time without abrupt or unnatural temporal changes.
\textit{Aesthetic Quality} evaluates the overall visual appeal of the generated content.
The reported VBench scores are obtained directly from the official benchmark evaluators, and higher values indicate better performance.

\paragraph{VBenchLong Metrics.}
For long-video generation, we employ VBenchLong~\citep{huang2025vbench++} to evaluate 30-second and 60-second videos. The main-paper figure reports overall quality, dynamic degree, and chunk discontinuity; Appendix~\ref{app:full_benchmark_results} also provides the available imaging quality, subject consistency, background consistency, motion smoothness, and aesthetic quality measurements. The overall score is aggregated using the standard VBench weighting coefficients. Compared with short-video evaluation, these metrics additionally reflect the ability of a model to maintain visual fidelity, subject identity, scene consistency, and temporal coherence over substantially longer autoregressive rollouts.

\paragraph{VisionReward.}
We further employ VisionReward~\citep{xu2026visionreward} as a complementary human-preference-based evaluator. Following the evaluation protocol of Causal Forcing, we evaluate generated videos on a 100-prompt subset and report both the overall VisionReward score and the instruction-following score. The overall reward measures the preference-aligned quality of the generated video, while the instruction-following score evaluates how faithfully the generated content satisfies the input text instruction. Higher values indicate better performance for both metrics.

\section{Details of Semantic Constraints and Prompt Refinement}
\label{sec:semantic_plan_fields}

\paragraph{Plan construction.}
For each unique benchmark prompt, we prepare one plan and keep it fixed during candidate generation. Given the original prompt $\mathbf{q}^{0}=\mathbf{q}$, the semantic planner returns the semantic constraints $\mathcal{R}$, prompt-specific exclusions $\mathcal{N}$, and refined prompt $\mathbf{q}^{1}$ in a single call, as in Eq.~\ref{eq:semantic_planning}. The original prompt remains unchanged.

The semantic constraints $\mathcal{R}$ list the explicit content to preserve, including subjects, counts, attributes, relations, actions, scenes, temporal cues, camera cues, styles, and the requested degree of motion. The prompt-specific exclusions $\mathcal{N}$ identify omissions, contradictions, and unrequested additions to avoid. The planner is instructed to rewrite $\mathbf{q}^{0}$ as a clear description that preserves the content in $\mathcal{R}$ while avoiding the changes identified by $\mathcal{N}$. It may clarify wording and composition, but must not introduce new subjects, objects, actions, scene changes, on-screen text, viewpoints, or camera motion. Explicit requests for stillness must be preserved, and a prompt that needs no refinement may remain unchanged. The three outputs are produced together by the planner; the two sets guide the rewrite rather than being concatenated to form $\mathbf{q}^{1}$. In particular, $\mathcal{N}$ is not supplied to the generator as a separate negative prompt. Table~\ref{tab:semantic_plan_fields} illustrates how a fragmented request can be rewritten without adding new content.

\begin{table*}[!htbp]
\centering
\small
\setlength{\tabcolsep}{6pt}
\renewcommand{\arraystretch}{1.15}
\caption{An illustrative example of conservative semantic planning. Matching colors trace the requested content from $\mathbf{q}^{0}$ through $\mathcal{R}$ to $\mathbf{q}^{1}$: \planentity{subjects and attributes}, \planaction{actions and their order}, and \plansetting{scene and camera}. \planavoid{Prompt-specific exclusions} identify changes to avoid.}
\label{tab:semantic_plan_fields}
\begin{tabularx}{\textwidth}{@{}>{\raggedright\arraybackslash}p{0.23\textwidth} >{\raggedright\arraybackslash}X@{}}
\toprule
\textbf{Component and role} & \textbf{Example} \\
\midrule

\textbf{Original prompt} $\mathbf{q}^{0}$\par Before refinement. &
``\plansetting{park}, \planentity{one girl one boy}, \planentity{red coat on girl blue coat on boy}, \planentity{yellow ball}, \planaction{girl gives it to boy then boy lifts it over his head}, \plansetting{fixed camera}'' \\
\midrule

\textbf{Semantic constraints} $\mathcal{R}$\par What to preserve. &
\planentity{$r_1$: One girl in a red coat.}\par
\planentity{$r_2$: One boy in a blue coat.}\par
\planentity{$r_3$: A yellow ball.}\par
\planaction{$r_4$: The girl gives the ball to the boy.}\par
\planaction{$r_5$: The boy then lifts the ball over his head.}\par
\plansetting{$r_6$: A park.}\par
\plansetting{$r_7$: A fixed camera.} \\
\midrule

\textbf{Prompt-specific exclusions} $\mathcal{N}$\par What to avoid. &
\planavoid{Swapped coat colors; reversed giver and receiver; reversed action order; camera movement; additional people, objects, or actions.} \\
\midrule

\textbf{Refined prompt} $\mathbf{q}^{1}$\par After refinement. &
``\plansetting{In a park}, \planentity{one girl in a red coat} \planaction{gives} \planentity{a yellow ball} \planaction{to} \planentity{one boy in a blue coat}. \planaction{The boy \textbf{then} lifts the \textbf{same ball} over his head}. \plansetting{The camera remains fixed}.'' \\
\bottomrule
\end{tabularx}

\vspace{3pt}
\begin{minipage}{\textwidth}
\end{minipage}
\end{table*}

\paragraph{Output checks and fallback.}
A returned plan contains one to ten semantic constraints and at most six prompt-specific exclusions. The planner is asked to use at most 55 English words for the refined prompt; the acceptance checks require a nonempty prompt of at most 700 characters and reject prompts exceeding 70 words. A short scenario label and a brief explanation of how the refinement preserves the request are also recorded. These checks verify output structure and length, rather than establish semantic equivalence. If the planning call or these checks fail during plan preparation, we use $\mathcal{R}=\{\mathbf{q}^{0}\}$, $\mathcal{N}=\varnothing$, and $\mathbf{q}^{1}=\mathbf{q}^{0}$. This fallback preserves the original request as one constraint rather than decomposing it into individual constraints. Both candidates are subsequently assessed against the original request by the vision--language critic.

\section{Theoretical Justification of Semantic KV Retention}
\label{app:semantic_theory}

This section proves Theorems~\ref{thm:semantic_mass_error}--\ref{thm:semantic_topk_bound} and explains their scope. The analysis follows the existing mean--maximum score and top-$k$ policy; it introduces no additional query sampling, optimization stage, or error threshold.

\paragraph{Setting and scope.}
Fix one candidate, one completed-chunk update $t$, and one layer/KV-head unit. The available cache $\mathcal{H}_t$, observed queries, native masks, and nonempty groups $\{\mathcal{P}_a\}_{a\in\mathcal{I}}$ are those used for scoring in Section~\ref{sec:budgeted_selection}, with $\mathcal{I}\neq\varnothing$. Queries and keys are held fixed when comparing attention before and after retention. A retained set $\mathcal{S}\subseteq\mathcal{H}_t$ contains the mandatory states $\mathcal{A}_t$. Whenever an output after retention is considered, $\mathcal{S}$ must contain at least one visible entry for every selected query; we call such a set \emph{usable}. Native mandatory states can ensure this condition, but the score alone does not. The reference is the currently available cache, not a reconstructed full-history cache. The discarded-mass bounds remain defined for every subset, even when its retained attention output would be undefined.

\subsection{From Discarded Attention to Constraint-Wise Output Error}
\label{app:semantic_error_bound}

\noindent\textit{Proof of Theorem~\ref{thm:semantic_mass_error}.}
For query $p$, define the discarded attention mass $m_p(\mathcal{S})=\sum_{j\in\mathcal{H}_t\setminus\mathcal{S}}\alpha_{pj}$. Masked entries have zero weight, and the visible weights sum to one, so $0\leq m_p(\mathcal{S})<1$ for a usable retained set. Because the retained keys and their logits are unchanged, the original and retained attention outputs are
\begin{equation}
\mathbf{o}_p=\sum_{j\in\mathcal{H}_t}\alpha_{pj}\mathbf{v}_j,
\qquad
\mathbf{o}_p^{\mathcal{S}}=\frac{\sum_{j\in\mathcal{S}}\alpha_{pj}\mathbf{v}_j}{1-m_p(\mathcal{S})}.
\label{eq:theory_attention_outputs}
\end{equation}
If $m_p(\mathcal{S})=0$, the two outputs are identical. Otherwise, let $\mathbf{o}_p^{\mathrm{drop}}=m_p(\mathcal{S})^{-1}\sum_{j\in\mathcal{H}_t\setminus\mathcal{S}}\alpha_{pj}\mathbf{v}_j$ be the normalized average of the discarded values. Splitting the original output gives
\begin{equation}
\mathbf{o}_p=(1-m_p(\mathcal{S}))\mathbf{o}_p^{\mathcal{S}}+m_p(\mathcal{S})\mathbf{o}_p^{\mathrm{drop}},
\qquad
\|\mathbf{o}_p-\mathbf{o}_p^{\mathcal{S}}\|_2
\leq D_{\mathrm{V}}\,m_p(\mathcal{S}).
\label{eq:theory_mixture_bound}
\end{equation}
To obtain the inequality, note that both normalized outputs are convex combinations of available values. Writing their difference as a convex combination of pairwise value differences and applying the triangle inequality bounds their distance by $D_{\mathrm{V}}$. Averaging the discarded mass within each group yields
\begin{equation}
\frac{1}{|\mathcal{P}_a|}\sum_{p\in\mathcal{P}_a}m_p(\mathcal{S})
=\sum_{j\in\mathcal{H}_t\setminus\mathcal{S}}\frac{1}{|\mathcal{P}_a|}\sum_{p\in\mathcal{P}_a}\alpha_{pj}
=\sum_{j\in\mathcal{H}_t\setminus\mathcal{S}}w_{aj}
=\ell_a(\mathcal{S}).
\label{eq:theory_mass_average}
\end{equation}
Averaging Eq.~\ref{eq:theory_mixture_bound} and then taking the largest groupwise error proves $\mathcal{E}(\mathcal{S};\mathbf{V}_t)\leq D_{\mathrm{V}}\mathcal{L}(\mathcal{S})$. \hfill$\square$

\paragraph{Sharpness and interpretation.}
For fixed queries, keys, masks, groups, and a usable $\mathcal{S}$, the bound is sharp over value tables whose diameter is at most a prescribed $D\geq0$:
\begin{equation}
\sup_{\mathbf{V}_t:\,\operatorname{diam}(\mathbf{V}_t)\leq D}\mathcal{E}(\mathcal{S};\mathbf{V}_t)
=D\,\mathcal{L}(\mathcal{S}).
\label{eq:theory_sharp_value_bound}
\end{equation}
Here, $\operatorname{diam}(\mathbf{V}_t)$ is the largest pairwise distance between the value vectors. The theorem gives the upper bound. For equality, choose a unit vector $\mathbf{u}$, assign $\mathbf{v}_j=\mathbf{0}$ to every retained entry, and assign $\mathbf{v}_j=D\mathbf{u}$ to every discarded entry. Then $\mathbf{o}_p^{\mathcal{S}}=\mathbf{0}$ and $\mathbf{o}_p=D\,m_p(\mathcal{S})\mathbf{u}$ for every query. This single shared value table attains equality for all groups simultaneously. The construction is a mathematical worst case with queries and keys frozen, not a claim that arbitrary value tables occur in the generator. Actual errors can be smaller through cancellation. No small-deletion approximation is used, and these outputs are measured before the attention output projection.

\subsection{Why the Maximum Score Leads to One \texorpdfstring{Top-$k$}{Top-k} Selection}
\label{app:semantic_topk_theory}

\noindent\textit{Proof of Theorem~\ref{thm:semantic_additive_envelope}.}
Recall the additive bound formed from the mean--maximum scores:
\begin{equation}
\mathcal{U}(\mathcal{S})=\sum_{j\in\mathcal{H}_t\setminus\mathcal{S}}s_j
=\sum_{j\in\mathcal{H}_t\setminus\mathcal{S}}\max_{a\in\mathcal{I}}w_{aj}.
\label{eq:theory_additive_envelope}
\end{equation}
Because the attention weights are nonnegative, $w_{aj}\leq s_j\leq\sum_{a'\in\mathcal{I}}w_{a'j}$ for every $a$ and $j$. Summing over the discarded entries gives
\begin{equation}
\mathcal{L}(\mathcal{S})=\max_a\ell_a(\mathcal{S})
\leq\mathcal{U}(\mathcal{S})
\leq\sum_{a\in\mathcal{I}}\ell_a(\mathcal{S})
\leq|\mathcal{I}|\,\mathcal{L}(\mathcal{S}).
\label{eq:theory_envelope_proof}
\end{equation}
To establish minimality, consider fixed nonnegative coefficients $\{b_j\}$ satisfying
\begin{equation}
\sum_{j\in\mathcal{H}_t\setminus\mathcal{S}}b_j\geq\mathcal{L}(\mathcal{S})
\quad\text{for every }\mathcal{A}_t\subseteq\mathcal{S}\subseteq\mathcal{H}_t.
\label{eq:theory_minimal_coefficients}
\end{equation}
For any optional entry $j$, take $\mathcal{S}=\mathcal{H}_t\setminus\{j\}$. The bound then requires $b_j\geq\max_a w_{aj}=s_j$. Conversely, $b_j\geq s_j$ for every optional entry is sufficient by Eq.~\ref{eq:theory_envelope_proof}. Thus, these coefficient-wise conditions are necessary and sufficient, proving that $\mathcal{U}$ is the pointwise smallest zero-intercept additive bound valid for every deletion of optional entries. \hfill$\square$

The universal requirement is important: it covers all deletion sets, not only the subsets satisfying one particular budget or mask-coverage condition. The singleton argument concerns discarded masses, which remain defined even if removing that entry leaves some query without a usable context. The theorem does not identify the tightest possible bound among nonadditive functions or bounds allowed to include an extra constant.

\medskip
\noindent\textit{Proof of Theorem~\ref{thm:semantic_topk_bound}.}
The count-based feasible family in the main text is
\begin{equation}
\mathcal{F}_t=\left\{\mathcal{S}\subseteq\mathcal{H}_t:\mathcal{A}_t\subseteq\mathcal{S},\ |\mathcal{S}\setminus\mathcal{A}_t|\leq k_t\right\}.
\label{eq:theory_feasible_family}
\end{equation}
For any $\mathcal{S}\in\mathcal{F}_t$, mandatory entries are never discarded, so
\begin{equation}
\mathcal{U}(\mathcal{S})
=\sum_{j\in\mathcal{H}_t\setminus\mathcal{A}_t}s_j
-\sum_{j\in\mathcal{S}\setminus\mathcal{A}_t}s_j.
\label{eq:theory_topk_complement}
\end{equation}
The first sum is constant. Since the scores are nonnegative and optional entries have equal costs, retaining the $k_t$ highest-scoring optional entries maximizes the second sum and therefore minimizes $\mathcal{U}$. If fewer optional entries are available, all are retained. Ties do not change the objective value. Let $\mathcal{S}^{\star}$ minimize $\mathcal{L}$ over the finite family $\mathcal{F}_t$. Theorem~\ref{thm:semantic_additive_envelope} and the established optimality imply
\begin{equation}
\mathcal{L}(\mathcal{S}_t)\leq\mathcal{U}(\mathcal{S}_t)
\leq\mathcal{U}(\mathcal{S}^{\star})
\leq|\mathcal{I}|\,\mathcal{L}(\mathcal{S}^{\star})
=|\mathcal{I}|\min_{\mathcal{S}\in\mathcal{F}_t}\mathcal{L}(\mathcal{S}).
\label{eq:theory_group_factor}
\end{equation}
This proves both statements. \hfill$\square$

\paragraph{What is optimized, and what is relaxed.}
If the returned set is usable, Theorems~\ref{thm:semantic_mass_error} and~\ref{thm:semantic_additive_envelope} also imply
\begin{equation}
\mathcal{E}(\mathcal{S}_t;\mathbf{V}_t)
\leq D_{\mathrm{V}}\mathcal{L}(\mathcal{S}_t)
\leq D_{\mathrm{V}}\mathcal{U}(\mathcal{S}_t)
=D_{\mathrm{V}}\min_{\mathcal{S}\in\mathcal{F}_t}\mathcal{U}(\mathcal{S}).
\label{eq:theory_topk_error}
\end{equation}
The exact optimization guarantee concerns $\mathcal{U}$, not actual output error or exact minimization of $\mathcal{L}$. For one group, $\mathcal{U}=\mathcal{L}$; with multiple groups, the relaxation can be conservative. Equal-cost, count-based feasibility is essential: arbitrary subset-dependent packing costs or additional mask-coverage constraints need separate treatment. If the selected top-$k$ set is itself usable, its upper-bound optimality also holds against the usable sets in $\mathcal{F}_t$, since these form a subset of the comparison family. The guarantees apply within a layer/KV-head unit under its assigned budget; they do not optimize the allocation across units. The value diameter is used only for analysis and need not be computed during retention.

\subsection{Why Keep Semantic Groups Separate?}
\label{app:semantic_grouping_theory}

Within-group averaging makes the score invariant to uniformly replicating all observations in any one group. Taking a maximum makes it invariant to adding a duplicate group with the same mean attention profile. Both statements follow directly from Eq.~\ref{eq:semantic_score}; arbitrary new observations need not leave the score unchanged. In contrast, pooling the observations counted with their group memberships produces
\begin{equation}
\begin{aligned}
s_j^{\mathrm{pool}}&=\sum_{a\in\mathcal{I}}\pi_a w_{aj},
\qquad
\pi_a=\frac{|\mathcal{P}_a|}{\sum_{b\in\mathcal{I}}|\mathcal{P}_b|},\\
\sum_{j\in\mathcal{H}_t\setminus\mathcal{S}}s_j^{\mathrm{pool}}
&=\sum_{a\in\mathcal{I}}\pi_a\ell_a(\mathcal{S}).
\end{aligned}
\label{eq:theory_pooled_groups}
\end{equation}
Thus, pooling controls a query-count-weighted average rather than a bound applying separately to every group. An entry receiving weight $w_{aj}$ only from group $a$ receives pooled score $\pi_a w_{aj}$, whereas its mean--maximum score remains $w_{aj}$. Even with equal group sizes, this attenuation is a factor of $|\mathcal{I}|$. More generally, with $\pi_{\min}=\min_a\pi_a>0$, the pooled discarded mass can upper-bound $\mathcal{L}$ only after a worst-case rescaling by $1/\pi_{\min}$; this factor is necessary when the discarded mass is concentrated in the least-weighted group. These observations explain the chosen aggregation, but do not imply that its retained set dominates pooled selection on every input.

\paragraph{Relation to prior work and practical meaning.}
Attention-output bounds and top-$k$ bound minimization have precedents in cache eviction~\citep{feng2026ada}. Mean--maximum aggregation across attention heads is also described in the OmniKV discussion of \citet{hao2026deltakv}. Here, the groups come from planner-extracted semantic constraints, and the analysis identifies the smallest universal additive bound for their individual discarded masses. Unlike the chunk-wise denoising analysis of Light Forcing~\citep{lv2026light}, the result requires neither a diffusion convergence rate nor a model of how sparsity changes score-estimation error. It is a local guarantee for the observed groups and unchanged KV entries. Whether cross-attention selects semantically informative queries, whether those observations represent later queries, and whether the retained information improves video quality remain empirical questions. No error tolerance, extra optimization term, or output-reconstruction module is added to \method{}.

\section{Detailed Benchmark Results}
\label{app:full_benchmark_results}

This section collects detailed numerical comparisons. Table~\ref{tab:main_result_backbone} retains all 16 VBench dimensions, throughput, speedup, and chunk discontinuity across backbones. The main-paper backbone table is a direct subset, with a common semantic diagnostic panel. For the compression-ratio comparison, Table~\ref{tab:main_result_short_full} reports the six dimensions in Figure~\ref{fig:cache_setting_radar}, together with throughput and speedup. Table~\ref{tab:main_result_long} compares \method{} with Full KV for 30- and 60-second generation; Table~\ref{tab:visionreward} reports the 100-prompt preference evaluation.

\subsection{VBench across Backbones}
\label{app:vbench_backbones}

Table~\ref{tab:main_result_backbone} expands the Self Forcing and Causal Forcing comparison in Table~\ref{tab:main_result_short}. Panel A reports efficiency and visual/temporal quality, and Panel B provides all nine semantic dimensions, including the six content diagnostics displayed in the main table.

\begin{table*}[htbp]
\centering
\captionsetup{skip=2pt}
\fontsize{8.0}{8.4}\selectfont
\setlength{\tabcolsep}{3.5pt}
\renewcommand{\arraystretch}{1.15}

\caption{Efficiency and generation quality on VBench for 5-second video generation at compression ratio $0.7$. Full KV retains the native cache without additional eviction and serves as the reference. Speedup is measured relative to Full KV. VBench metrics retain their reported percentage-scaled values. $\uparrow$ indicates higher is better and $\downarrow$ indicates lower is better. \textbf{Bold} marks the best result among compressed methods for each backbone.}
\label{tab:main_result_backbone}

\textbf{Panel A: Efficiency, Temporal Quality, and Overall Video Quality}

\vspace{2pt}

\resizebox{\textwidth}{!}{%
\begin{tabular}{@{}lllcc*{2}{c}*{4}{c}*{5}{c}@{}}
\toprule
\multirow{2}{*}{\textbf{Backbone}}
& \multirow{2}{*}{\textbf{Length}}
& \multirow{2}{*}{\textbf{Method}}
& \multicolumn{2}{c}{\textbf{Efficiency}}
& \multicolumn{2}{c}{\textbf{KV Cache}}
& \multicolumn{4}{c}{\textbf{Temporal Quality}}
& \multicolumn{5}{c}{\textbf{VBench Quality}} \\
\cmidrule(lr){4-5}\cmidrule(lr){6-7}\cmidrule(lr){8-11}\cmidrule(lr){12-16}
& & &
\textbf{FPS}$\uparrow$
& \textbf{Speedup}$\uparrow$
& \shortstack{\textbf{Memory}\\\textbf{(GB)}$\downarrow$}
& \shortstack{\textbf{Reduction}\\\textbf{($\times$)}$\uparrow$}
& \shortstack{\textbf{Chunk}\\\textbf{Disc.}$\downarrow$}
& \shortstack{\textbf{Dynamic}\\\textbf{Degree}$\uparrow$}
& \shortstack{\textbf{Temporal}\\\textbf{Flicker.}$\uparrow$}
& \shortstack{\textbf{Motion}\\\textbf{Smooth.}$\uparrow$}
& \shortstack{\textbf{Total}\\\textbf{Score}$\uparrow$}
& \shortstack{\textbf{Imaging}\\\textbf{Quality}$\uparrow$}
& \shortstack{\textbf{Subject}\\\textbf{Cons.}$\uparrow$}
& \shortstack{\textbf{Background}\\\textbf{Cons.}$\uparrow$}
& \shortstack{\textbf{Aesthetic}\\\textbf{Quality}$\uparrow$} \\
\midrule

\multirow{5}{*}{Self Forcing}
& \multirow{5}{*}{5 s}
& \fullkvcell{Full KV}
& \fullkvcell{6.95}
& \fullkvcell{1.00$\times$}
& \fullkvcell{6.04} & \fullkvcell{1.00$\times$}
& \fullkvcell{2.25}
& \fullkvcell{91.67}
& \fullkvcell{98.54}
& \fullkvcell{97.77}
& \fullkvcell{82.96}
& \fullkvcell{67.50}
& \fullkvcell{93.84}
& \fullkvcell{95.24}
& \fullkvcell{61.07} \\

\cdashline{3-16}

& & R-KV
& 7.36
& 1.06$\times$
& 1.81 & 3.35$\times$
& \textbf{2.23}
& 92.78
& 98.45
& 97.77
& 82.90
& 67.52
& \textbf{93.54}
& 95.01
& \textbf{61.26} \\

& & Dummy Forcing
& \textbf{8.16}
& \textbf{1.17$\times$}
& 1.81 & 3.35$\times$
& \textbf{2.23}
& 91.39
& 98.42
& \textbf{97.81}
& 82.87
& 67.49
& 93.49
& \textbf{95.22}
& 61.09 \\

& & Forcing-KV
& 7.92
& 1.14$\times$
& 1.81 & 3.35$\times$
& 2.62
& 92.78
& 98.41
& 97.55
& 82.69
& 67.06
& 93.28
& 95.14
& 60.95 \\

\rowcolor{softcyan}
& & \textbf{\method{}}
& 7.95
& 1.14$\times$
& 1.81 & 3.35$\times$
& 2.26
& \textbf{93.33}
& \textbf{98.50}
& 97.80
& \textbf{83.03}
& \textbf{67.56}
& 93.49
& 95.10
& 60.99 \\

\midrule

\multirow{5}{*}{Causal Forcing}
& \multirow{5}{*}{5 s}
& \fullkvcell{Full KV}
& \fullkvcell{7.24}
& \fullkvcell{1.00$\times$}
& \fullkvcell{6.04} & \fullkvcell{1.00$\times$}
& \fullkvcell{2.39}
& \fullkvcell{83.61}
& \fullkvcell{98.44}
& \fullkvcell{97.87}
& \fullkvcell{83.99}
& \fullkvcell{71.12}
& \fullkvcell{96.62}
& \fullkvcell{97.07}
& \fullkvcell{64.03} \\

\cdashline{3-16}

& & R-KV
& 8.14
& 1.13$\times$
& 1.81 & 3.35$\times$
& 2.54
& 86.39
& 98.32
& 97.72
& 84.00
& 71.17
& 96.32
& 96.81
& 64.24 \\

& & Dummy Forcing
& \textbf{9.20}
& \textbf{1.27$\times$}
& 1.81 & 3.35$\times$
& 2.53
& 86.39
& \textbf{98.35}
& \textbf{97.78}
& 84.05
& \textbf{71.21}
& \textbf{96.42}
& 96.87
& \textbf{64.33} \\

& & Forcing-KV
& 8.91
& 1.23$\times$
& 1.81 & 3.35$\times$
& 2.97
& \textbf{90.00}
& 98.23
& 97.39
& 84.04
& 70.89
& 96.14
& 96.96
& 64.03 \\

\rowcolor{softcyan}
& & \textbf{\method{}}
& 8.83
& 1.22$\times$
& 1.81 & 3.35$\times$
& \textbf{2.52}
& 86.67
& 98.34
& 97.72
& \textbf{84.10}
& 71.13
& 96.33
& \textbf{96.97}
& 64.13 \\

\bottomrule
\end{tabular}%
}

\vspace{7pt}

\textbf{Panel B: Semantic Fidelity}

\vspace{2pt}

\resizebox{\textwidth}{!}{%
\begin{tabular}{@{}lll*{9}{c}@{}}
\toprule
\textbf{Backbone}
& \textbf{Length}
& \textbf{Method}
& \shortstack{\textbf{Object}\\\textbf{Class}$\uparrow$}
& \shortstack{\textbf{Multiple}\\\textbf{Objects}$\uparrow$}
& \shortstack{\textbf{Human}\\\textbf{Action}$\uparrow$}
& \textbf{Color}$\uparrow$
& \shortstack{\textbf{Spatial}\\\textbf{Relation}$\uparrow$}
& \textbf{Scene}$\uparrow$
& \shortstack{\textbf{Temporal}\\\textbf{Style}$\uparrow$}
& \shortstack{\textbf{Appearance}\\\textbf{Style}$\uparrow$}
& \shortstack{\textbf{Overall}\\\textbf{Cons.}$\uparrow$} \\
\midrule

\multirow{5}{*}{Self Forcing}
& \multirow{5}{*}{5 s}
& \fullkvcell{Full KV}
& \fullkvcell{89.83}
& \fullkvcell{75.55}
& \fullkvcell{92.00}
& \fullkvcell{90.54}
& \fullkvcell{71.67}
& \fullkvcell{48.05}
& \fullkvcell{24.43}
& \fullkvcell{19.85}
& \fullkvcell{26.31} \\

\cdashline{3-12}

& & R-KV
& 88.29
& 74.15
& \textbf{93.00}
& \textbf{91.40}
& 71.66
& 47.53
& 24.35
& 19.85
& 26.30 \\

& & Dummy Forcing
& 89.26
& \textbf{77.07}
& 92.00
& 91.12
& 71.01
& 47.43
& 24.38
& 19.85
& \textbf{26.40} \\

& & Forcing-KV
& 88.21
& 75.88
& 92.60
& 90.64
& 71.44
& 47.72
& 24.42
& 19.76
& 26.39 \\

\rowcolor{softcyan}
& & \textbf{\method{}}
& \textbf{89.43}
& 75.47
& 92.00
& 91.30
& \textbf{72.46}
& \textbf{48.65}
& \textbf{24.45}
& \textbf{19.92}
& 26.37 \\

\midrule

\multirow{5}{*}{Causal Forcing}
& \multirow{5}{*}{5 s}
& \fullkvcell{Full KV}
& \fullkvcell{91.11}
& \fullkvcell{75.88}
& \fullkvcell{90.00}
& \fullkvcell{88.72}
& \fullkvcell{77.57}
& \fullkvcell{46.25}
& \fullkvcell{24.51}
& \fullkvcell{20.05}
& \fullkvcell{25.54} \\

\cdashline{3-12}

& & R-KV
& \textbf{92.88}
& 76.77
& 91.40
& 87.89
& 75.60
& 45.20
& 24.48
& 19.98
& 25.55 \\

& & Dummy Forcing
& 91.72
& 76.34
& \textbf{91.60}
& 88.12
& 77.57
& 43.88
& 24.52
& \textbf{20.07}
& 25.30 \\

& & Forcing-KV
& 92.74
& \textbf{79.19}
& 89.80
& \textbf{89.27}
& \textbf{78.18}
& 43.18
& \textbf{24.53}
& 19.97
& \textbf{25.60} \\

\rowcolor{softcyan}
& & \textbf{\method{}}
& 91.85
& 77.09
& \textbf{91.60}
& 89.09
& 78.01
& \textbf{45.35}
& 24.40
& 20.04
& 25.56 \\

\bottomrule
\end{tabular}%
}

\end{table*}

\subsection{VBench across Compression Ratios}
\label{app:vbench_cache_settings}

Table~\ref{tab:main_result_short_full} gives the numerical results underlying Figure~\ref{fig:cache_setting_radar} for compression ratios $0.5$, $0.7$, and $0.9$ on LongLive model. It retains Subject Consistency, Background Consistency, Temporal Flickering, Human Action, Color, and Object Class, together with DiT-only throughput and speedup. Full KV is shared only within this run group. The labels denote compression ratios, i.e., target fractions of the cache removed, rather than measured per-update byte reductions or peak-memory reductions.

\begin{table*}[htbp]
\centering
\captionsetup{skip=2pt}
\fontsize{8.0}{8.4}\selectfont
\setlength{\tabcolsep}{3.5pt}
\renewcommand{\arraystretch}{1.15}

\caption{VBench results across compression ratios for 5-second generation, reporting the six dimensions in Figure~\ref{fig:cache_setting_radar} and DiT-only efficiency. Full KV is the shared uncompressed reference within this run group. Speedup is measured relative to Full KV. VBench metrics retain their reported percentage-scaled values. $\uparrow$ indicates higher is better. \textbf{Bold} marks the best result among compressed methods at each compression ratio.}
\label{tab:main_result_short_full}

\resizebox{\textwidth}{!}{%
\begin{tabular}{@{}lllcc*{2}{c}*{6}{c}@{}}
\toprule
\multirow{2}{*}{\shortstack{\textbf{Compression}\\\textbf{Ratio}}}
& \multirow{2}{*}{\textbf{Length}}
& \multirow{2}{*}{\textbf{Method}}
& \multicolumn{2}{c}{\textbf{Efficiency}}
& \multicolumn{2}{c}{\textbf{KV Cache}}
& \multicolumn{6}{c}{\textbf{VBench Quality}} \\
\cmidrule(lr){4-5}\cmidrule(lr){6-7}\cmidrule(lr){8-13}
& & &
\textbf{FPS}$\uparrow$
& \textbf{Speedup}$\uparrow$
& \shortstack{\textbf{Memory}\\\textbf{(GB)}$\downarrow$}
& \shortstack{\textbf{Reduction}\\\textbf{($\times$)}$\uparrow$}
& \shortstack{\textbf{Temporal}\\\textbf{Flicker.}$\uparrow$}
& \shortstack{\textbf{Subject}\\\textbf{Cons.}$\uparrow$}
& \shortstack{\textbf{Background}\\\textbf{Cons.}$\uparrow$}
& \shortstack{\textbf{Object}\\\textbf{Class}$\uparrow$}
& \shortstack{\textbf{Human}\\\textbf{Action}$\uparrow$}
& \textbf{Color}$\uparrow$ \\
\midrule
\fullkvcell{} & \fullkvcell{} & \fullkvcell{Full KV} & \fullkvcell{6.57} & \fullkvcell{1.00$\times$} & \fullkvcell{3.45} & \fullkvcell{1.00$\times$} & \fullkvcell{99.26} & \fullkvcell{96.06} & \fullkvcell{96.68} & \fullkvcell{90.84} & \fullkvcell{92.60} & \fullkvcell{90.34} \\

\cdashline{1-13}

\multirow{4}{*}{0.5} & \multirow{4}{*}{5 s} & R-KV & 7.53 & 1.15$\times$ & 1.72 & 2.01$\times$ & \textbf{99.32} & \textbf{96.07} & 96.61 & 91.01 & 93.00 & \textbf{90.84} \\

 &  & Dummy Forcing & 7.32 & 1.11$\times$ & 1.72 & 2.01$\times$ & 98.81 & 95.61 & 95.95 & 90.09 & 93.60 & 88.95 \\

 &  & Forcing-KV & 7.67 & 1.17$\times$ & 1.72 & 2.01$\times$ & 99.29 & 95.92 & \textbf{96.63} & 90.84 & \textbf{94.00} & 89.73 \\

\rowcolor{softcyan}
 &  & \textbf{\method{}} & \textbf{8.00} & \textbf{1.22$\times$} & 1.72 & 2.01$\times$ & 99.29 & 96.04 & 96.52 & \textbf{91.93} & 92.80 & 89.24 \\

\midrule

\multirow{4}{*}{0.7} & \multirow{4}{*}{5 s} & R-KV & 8.47 & 1.29$\times$ & 1.02 & 3.38$\times$ & 99.24 & 95.88 & 96.37 & 91.01 & 92.80 & 88.10 \\

 &  & Dummy Forcing & 7.82 & 1.19$\times$ & 1.02 & 3.38$\times$ & 99.01 & 95.35 & 94.50 & \textbf{91.57} & 92.40 & 87.88 \\

 &  & Forcing-KV & 8.67 & \textbf{1.32$\times$} & 1.02 & 3.38$\times$ & 99.26 & 95.81 & \textbf{96.41} & 90.11 & \textbf{93.60} & 88.61 \\

\rowcolor{softcyan}
 &  & \textbf{\method{}} & \textbf{8.70} & \textbf{1.32$\times$} & 1.02 & 3.38$\times$ & \textbf{99.27} & \textbf{95.97} & \textbf{96.41} & 91.27 & 93.00 & \textbf{89.29} \\

\midrule

\multirow{4}{*}{0.9} & \multirow{4}{*}{5 s} & R-KV & \textbf{10.04} & \textbf{1.53$\times$} & 0.34 & 10.12$\times$ & 99.02 & 94.03 & 93.70 & 89.73 & 93.40 & \textbf{90.87} \\

 &  & Dummy Forcing & 8.84 & 1.34$\times$ & 0.34 & 10.12$\times$ & \textbf{99.04} & \textbf{94.48} & \textbf{94.17} & 88.78 & 92.60 & 90.15 \\

 &  & Forcing-KV & 9.95 & 1.51$\times$ & 0.34 & 10.12$\times$ & 98.79 & 93.70 & 92.99 & 87.91 & 93.80 & 90.29 \\

\rowcolor{softcyan}
 &  & \textbf{\method{}} & 10.00 & 1.52$\times$ & 0.34 & 10.12$\times$ & 99.02 & 94.07 & 93.70 & \textbf{91.60} & \textbf{94.60} & 90.69 \\
\bottomrule
\end{tabular}%
}
\end{table*}

\subsection{Long-Horizon Evaluation against Full KV}
\label{app:vbenchlong_full}

Table~\ref{tab:main_result_long} compares \method{} with Full KV on LongLive 2.0 for 30- and 60-second VBenchLong generation. This reference comparison measures the quality--throughput trade-off relative to uncompressed generation, with the same displayed metrics at both durations.

\paragraph{Thirty-second generation.}
\method{} increases DiT-only throughput from $4.86$ to $6.16$ FPS ($1.27\times$), while Total Score decreases from $80.44$ to $79.66$ ($0.78$ points). Subject and Background Consistency decrease by $0.40$ and $0.46$ points, respectively. Chunk discontinuity decreases from $3.41$ to $3.31$, and Dynamic Degree increases from $58.75$ to $60.40$. Imaging and Aesthetic Quality decrease from $66.89$ and $55.35$ to $64.40$ and $54.22$.

\paragraph{Sixty-second generation.}
\method{} increases throughput from $4.76$ to $6.12$ FPS ($1.29\times$), with Total Score decreasing from $80.12$ to $78.84$ ($1.28$ points). Subject and Background Consistency decrease by $0.46$ and $0.35$ points, while chunk discontinuity remains $3.39$. Imaging and Aesthetic Quality decrease by $3.10$ and $1.89$ points. The Total Score reduction relative to Full KV is larger at 60 seconds than at 30 seconds.

\paragraph{Scope of the evidence.}
These results characterize the quality--throughput trade-off relative to Full KV, not a ranking among compression methods. The local attention-output bounds concern observed query groups at a single cache update and do not guarantee unchanged video quality over long autoregressive rollouts.

\begin{table*}[t!]
\centering
\captionsetup{skip=2pt}
\small\selectfont
\setlength{\tabcolsep}{1.8pt}
\renewcommand{\arraystretch}{1.15}

\caption{
Comparison of \method{} with Full KV on VBenchLong for 30- and 60-second video generation.
Full KV retains the native cache without additional eviction and serves as the reference.
Speedup is measured relative to Full KV.
$\uparrow$ indicates higher is better and $\downarrow$ indicates lower is better.
The comparison measures the quality--throughput trade-off relative to the uncompressed reference; no ranking among compression methods is reported.
}
\label{tab:main_result_long}

\resizebox{\textwidth}{!}{%
\begin{tabular}{@{}lllcc*{2}{c}*{2}{c}*{6}{c}@{}}
\toprule

\multirow{2}{*}{\textbf{Backbone}}
& \multirow{2}{*}{\textbf{Length}}
& \multirow{2}{*}{\textbf{Method}}
& \multicolumn{2}{c}{\textbf{Efficiency}}
& \multicolumn{2}{c}{\textbf{KV Cache}}
& \multicolumn{2}{c}{\textbf{Temporal Quality}}
& \multicolumn{6}{c}{\textbf{VBenchLong Quality}} \\

\cmidrule(lr){4-5}
\cmidrule(lr){6-7}
\cmidrule(lr){8-9}
\cmidrule(lr){10-15}

&
&
&
\textbf{FPS}$\uparrow$
& \textbf{Speedup}$\uparrow$
& \shortstack{\textbf{Memory}\\\textbf{(GB)}$\downarrow$}
& \shortstack{\textbf{Reduction}\\\textbf{($\times$)}$\uparrow$}
& \shortstack{\textbf{Chunk}\\\textbf{Disc.}$\downarrow$}
& \shortstack{\textbf{Dynamic}\\\textbf{Degree}$\uparrow$}
& \shortstack{\textbf{Total}\\\textbf{Score}$\uparrow$}
& \shortstack{\textbf{Imaging}\\\textbf{Quality}$\uparrow$}
& \shortstack{\textbf{Subject}\\\textbf{Cons.}$\uparrow$}
& \shortstack{\textbf{Background}\\\textbf{Cons.}$\uparrow$}
& \shortstack{\textbf{Motion}\\\textbf{Smooth.}$\uparrow$}
& \shortstack{\textbf{Aesthetic}\\\textbf{Quality}$\uparrow$} \\

\midrule

\multirow{2}{*}{LongLive 2.0}
& \multirow{2}{*}{30 s}
& \fullkvcell{Full KV}
& \fullkvcell{4.86}
& \fullkvcell{1.00$\times$}
& \fullkvcell{20.76} & \fullkvcell{1.00$\times$}
& \fullkvcell{3.41}
& \fullkvcell{58.75}
& \fullkvcell{80.44}
& \fullkvcell{66.89}
& \fullkvcell{97.79}
& \fullkvcell{97.35}
& \fullkvcell{98.87}
& \fullkvcell{55.35} \\

\cdashline{3-15}

& & \cellcolor{softcyan}\textbf{\method{}}
& \cellcolor{softcyan}6.16 & \cellcolor{softcyan}1.27$\times$ & \cellcolor{softcyan}6.23 & \cellcolor{softcyan}3.33$\times$ & \cellcolor{softcyan}3.31 & \cellcolor{softcyan}60.40
& \cellcolor{softcyan}79.66 & \cellcolor{softcyan}64.40 & \cellcolor{softcyan}97.39 & \cellcolor{softcyan}96.89 & \cellcolor{softcyan}98.75 & \cellcolor{softcyan}54.22 \\
\midrule

\multirow{2}{*}{LongLive 2.0}
& \multirow{2}{*}{60 s}
& \fullkvcell{Full KV}
& \fullkvcell{4.76}
& \fullkvcell{1.00$\times$}
& \fullkvcell{20.76} & \fullkvcell{1.00$\times$}
& \fullkvcell{3.39}
& \fullkvcell{62.21}
& \fullkvcell{80.12}
& \fullkvcell{65.82}
& \fullkvcell{97.66}
& \fullkvcell{97.13}
& \fullkvcell{98.75}
& \fullkvcell{53.77} \\

\cdashline{3-15}

& & \cellcolor{softcyan}\textbf{\method{}}
& \cellcolor{softcyan}6.12 & \cellcolor{softcyan}1.29$\times$ & \cellcolor{softcyan}6.23 & \cellcolor{softcyan}3.33$\times$ & \cellcolor{softcyan}3.39 & \cellcolor{softcyan}61.15
& \cellcolor{softcyan}78.84 & \cellcolor{softcyan}62.72 & \cellcolor{softcyan}97.20 & \cellcolor{softcyan}96.78 & \cellcolor{softcyan}98.62 & \cellcolor{softcyan}51.88 \\

\bottomrule
\end{tabular}%
}
\end{table*}

\subsection{VisionReward Results}
\label{app:visionreward_full}

Table~\ref{tab:visionreward} reports overall VisionReward and instruction-following measurements on LongLive at compression ratio $0.9$. All methods are scored with the same VisionReward model and evaluator runtime on the separate 100-prompt set; these scores are not combined numerically with VBench or with online critic utilities. \method{} obtains $98.000$ instruction following, matching R-KV and Forcing-KV and exceeding Full KV and Dummy Forcing by two points. Its $7.964$ overall reward exceeds R-KV by $0.458$ and Forcing-KV by $0.181$, but trails Dummy Forcing by $0.497$. Thus, the method improves instruction following relative to some baselines without uniformly maximizing the broader reward aggregate.

\begin{table}[htbp]
\centering
\captionsetup{skip=2pt}
\small
\setlength{\tabcolsep}{6pt}
\renewcommand{\arraystretch}{1.15}
\caption{VisionReward evaluation on LongLive at compression ratio 0.9. We use the 100 prompts from Causal Forcing and score all methods with the same evaluator runtime. Higher is better for both metrics.}
\label{tab:visionreward}

\begin{tabular}{@{}lcc@{}}
\toprule
\textbf{Method}
& \textbf{VisionReward}$\uparrow$
& \textbf{Instruction Following}$\uparrow$ \\
\midrule

\fullkvcell{Full KV}
& \fullkvcell{10.264}
& \fullkvcell{96.000} \\

R-KV
& 7.506
& 98.000 \\

Dummy Forcing
& 8.461
& 96.000 \\

Forcing-KV
& 7.783
& 98.000 \\

\rowcolor{softcyan}
\textbf{\method{}}
& 7.964
& 98.000 \\

\bottomrule
\end{tabular}
\end{table}

\section{Detailed Ablation Results}
\label{app:ablation_studies}

The ablations use 5-second VBench generation with LongLive at compression ratio $0.9$. Prompts, seeds, cache budget, protected native states, generation schedule, and evaluation protocol are unchanged. The planner and complete agentic workflow remain active in every variant.

\subsection{Semantic Query Selection and Aggregation}
\label{app:ablation_efficiency}

With $n_q=32$, \emph{Uniform queries} replaces constraint-conditioned selection with the same number of uniformly spaced observations, preserving group sizes and maximum aggregation. \emph{Pooled constraints} keeps the selected queries but replaces the groupwise maximum in Eq.~\ref{eq:semantic_score} with pooled-mean attention. Each control changes only one component of cache scoring. We focus on whether these components improve semantic fidelity under the same cache budget. Table~\ref{tab:ablation_full} first reports the official VBench Semantic Score and its nine constituent dimensions, followed by efficiency, Quality Score, and temporal diagnostics.

\begin{table*}[htbp]
\centering
\captionsetup{skip=2pt}
\fontsize{8.0}{8.4}\selectfont
\setlength{\tabcolsep}{2pt}
\renewcommand{\arraystretch}{1.15}

\caption{Component ablations on LongLive and 5-second VBench at compression ratio $0.9$ with $n_q=32$. Panel A reports semantic fidelity; Panel B reports DiT-only efficiency and visual/temporal diagnostics. Semantic and Quality are the official VBench aggregate scores. VBench scores retain their reported percentage-scaled values. $\uparrow$ indicates higher is better and $\downarrow$ indicates lower is better. \textbf{Bold} marks the best displayed value among the three configurations.}
\label{tab:ablation_full}

\textbf{Panel A: Semantic Score and Semantic Fidelity}

\vspace{2pt}

\begin{tabular}{@{}l*{10}{c}@{}}
\toprule
\textbf{Configuration}
& \shortstack{\textbf{Semantic}\\\textbf{Score}$\uparrow$}
& \shortstack{\textbf{Object}\\\textbf{Class}$\uparrow$}
& \shortstack{\textbf{Multiple}\\\textbf{Objects}$\uparrow$}
& \shortstack{\textbf{Human}\\\textbf{Action}$\uparrow$}
& \textbf{Color}$\uparrow$
& \shortstack{\textbf{Spatial}\\\textbf{Relation}$\uparrow$}
& \textbf{Scene}$\uparrow$
& \shortstack{\textbf{Temporal}\\\textbf{Style}$\uparrow$}
& \shortstack{\textbf{Appearance}\\\textbf{Style}$\uparrow$}
& \shortstack{\textbf{Overall}\\\textbf{Cons.}$\uparrow$} \\
\midrule

Uniform queries
& 76.94 & 90.44 & \textbf{76.02} & \textbf{94.60}
& 90.26 & 72.12 & 44.36 & 25.05 & 20.61 & \textbf{27.00} \\

Pooled constraints
& 76.78 & 91.00 & 75.90 & 93.80
& 89.40 & 73.05 & 43.71 & 25.04 & 20.52 & 26.98 \\

\rowcolor{softcyan}
\textbf{\method{}}
& \textbf{77.18} & \textbf{91.60} & 75.34 & \textbf{94.60}
& \textbf{90.69} & \textbf{73.27} & \textbf{44.46}
& \textbf{25.09} & \textbf{20.66} & 26.88 \\

\bottomrule
\end{tabular}

\vspace{7pt}

\textbf{Panel B: Efficiency and Visual/Temporal Diagnostics}

\vspace{2pt}

\setlength{\tabcolsep}{4pt}
\begin{tabular}{@{}lcc*{2}{c}*{5}{c}@{}}
\toprule
\multirow{2}{*}{\textbf{Configuration}}
& \multicolumn{2}{c}{\textbf{Efficiency}}
& \multicolumn{2}{c}{\textbf{KV Cache}}
& \multicolumn{5}{c}{\textbf{Visual/Temporal Diagnostics}} \\
\cmidrule(lr){2-3}\cmidrule(lr){4-5}\cmidrule(lr){6-10}
& \textbf{FPS}$\uparrow$
& \textbf{Speedup}$\uparrow$
& \shortstack{\textbf{Memory}\\\textbf{(GB)}$\downarrow$}
& \shortstack{\textbf{Reduction}\\\textbf{($\times$)}$\uparrow$}
& \shortstack{\textbf{Quality}\\\textbf{Score}$\uparrow$}
& \shortstack{\textbf{Chunk}\\\textbf{Disc.}$\downarrow$}
& \shortstack{\textbf{Dynamic}\\\textbf{Degree}$\uparrow$}
& \shortstack{\textbf{Temporal}\\\textbf{Flicker.}$\uparrow$}
& \shortstack{\textbf{Motion}\\\textbf{Smooth.}$\uparrow$} \\
\midrule

Uniform queries
& 9.98 & \textbf{1.52$\times$} & 0.34 & 10.12$\times$ & \textbf{82.93}
& \textbf{3.45} & \textbf{62.50} & \textbf{99.02} & 98.55 \\

Pooled constraints
& 9.94 & 1.51$\times$ & 0.34 & 10.12$\times$ & 82.64
& 3.52 & 59.44 & 99.01 & 98.56 \\

\rowcolor{softcyan}
\textbf{\method{}}
& \textbf{10.00} & \textbf{1.52$\times$} & 0.34 & 10.12$\times$ & 82.80
& 3.54 & 61.39 & \textbf{99.02} & \textbf{98.58} \\

\bottomrule
\end{tabular}
\end{table*}

\paragraph{Semantic effects.}
Constraint-conditioned selection improves the Semantic Score over uniform queries by $0.24$ points, with gains of $1.16$ in Object Class and $1.15$ in Spatial Relation; Human Action is tied at $94.60$. Relative to pooled constraints, the full method improves the Semantic Score by $0.40$ points, including gains of $0.60$ in Object Class, $1.29$ in Color, and $0.75$ in Scene. These comparisons support the semantic contributions of query selection and groupwise maximum aggregation under the fixed cache budget.

\paragraph{Efficiency and scope.}
Throughput ranges from $9.94$ to $10.00$ FPS, so the measurements are similar across configurations rather than evidence of a substantial speed advantage. The full method does not dominate every metric: uniform queries achieve a higher Quality Score and lower chunk discontinuity, while both controls score higher on Multiple Objects and Overall Consistency. We interpret the results as a semantic-alignment benefit, not uniformly better visual or temporal quality, and do not attach statistical significance to the small score differences.

\subsection{Semantic Query Limit}
\label{sec:query_limit_ablation}

The query limit $n_q$ bounds the number of selected self-attention queries per semantic constraint; it does not change the target cache budget. We compare $n_q\in\{16,32,64\}$ with all other settings fixed. Table~\ref{tab:query_limit_ablation} retains the complete results.

\begin{table*}[htbp]
\centering
\captionsetup{skip=2pt}
\fontsize{8.2}{8.6}\selectfont
\setlength{\tabcolsep}{5.0pt}
\renewcommand{\arraystretch}{1.15}

\caption{Ablation of the semantic query limit $n_q$ on VBench. The compression ratio is fixed to $0.9$. VBench metrics retain their percentage-scaled values, except Chunk Discontinuity, which is the RAFT TopK-to-mean optical-flow ratio. $\uparrow$ indicates higher is better and $\downarrow$ indicates lower is better. The best displayed result in each column is highlighted in \textbf{bold}.}
\label{tab:query_limit_ablation}

\textbf{Panel A: Efficiency, Temporal Quality, and Overall Video Quality}

\vspace{2pt}

\resizebox{\textwidth}{!}{%
\begin{tabular}{@{}c cc cc cccc ccccc@{}}
\toprule
\multirow{2}{*}{$n_q$}
& \multicolumn{2}{c}{\textbf{Efficiency}}
& \multicolumn{2}{c}{\textbf{KV Cache}}
& \multicolumn{4}{c}{\textbf{Temporal Quality}}
& \multicolumn{5}{c}{\textbf{VBench Quality}} \\
\cmidrule(lr){2-3}\cmidrule(lr){4-5}\cmidrule(lr){6-9}\cmidrule(lr){10-14}
& \textbf{FPS}$\uparrow$
& \textbf{Speedup}$\uparrow$
& \shortstack{\textbf{Memory}\\\textbf{(GB)}$\downarrow$}
& \shortstack{\textbf{Reduction}\\\textbf{($\times$)}$\uparrow$}
& \shortstack{\textbf{Chunk}\\\textbf{Disc.}$\downarrow$}
& \shortstack{\textbf{Dynamic}\\\textbf{Degree}$\uparrow$}
& \shortstack{\textbf{Temporal}\\\textbf{Flicker.}$\uparrow$}
& \shortstack{\textbf{Motion}\\\textbf{Smooth.}$\uparrow$}
& \shortstack{\textbf{Total}\\\textbf{Score}$\uparrow$}
& \shortstack{\textbf{Imaging}\\\textbf{Quality}$\uparrow$}
& \shortstack{\textbf{Subject}\\\textbf{Cons.}$\uparrow$}
& \shortstack{\textbf{Background}\\\textbf{Cons.}$\uparrow$}
& \shortstack{\textbf{Aesthetic}\\\textbf{Quality}$\uparrow$} \\
\midrule
$16$ & 9.93 & 1.51$\times$ & 0.34 & 10.12$\times$ & \textbf{3.54} & 60.00 & \textbf{99.02} & 98.57 & 81.54 & 67.35 & 94.00 & 93.68 & \textbf{62.45} \\
\rowcolor{softcyan}
$\mathbf{32}$ & \textbf{10.00} & \textbf{1.52$\times$} & 0.34 & 10.12$\times$ & \textbf{3.54} & \textbf{61.39} & \textbf{99.02} & \textbf{98.58} & \textbf{81.68} & \textbf{67.36} & \textbf{94.07} & 93.70 & 62.30 \\
$64$ & 9.78 & 1.49$\times$ & 0.34 & 10.12$\times$ & 3.55 & 58.33 & \textbf{99.02} & 98.57 & 81.45 & 67.27 & 93.95 & \textbf{93.81} & 62.34 \\
\bottomrule
\end{tabular}%
}

\vspace{6pt}

\textbf{Panel B: Semantic Fidelity}

\vspace{2pt}

\resizebox{\textwidth}{!}{%
\begin{tabular}{@{}c ccccccccc@{}}
\toprule
$n_q$
& \shortstack{\textbf{Object}\\\textbf{Class}$\uparrow$}
& \shortstack{\textbf{Multiple}\\\textbf{Objects}$\uparrow$}
& \shortstack{\textbf{Human}\\\textbf{Action}$\uparrow$}
& \textbf{Color}$\uparrow$
& \shortstack{\textbf{Spatial}\\\textbf{Relation}$\uparrow$}
& \textbf{Scene}$\uparrow$
& \shortstack{\textbf{Temporal}\\\textbf{Style}$\uparrow$}
& \shortstack{\textbf{Appearance}\\\textbf{Style}$\uparrow$}
& \shortstack{\textbf{Overall}\\\textbf{Cons.}$\uparrow$} \\
\midrule
$16$ & 90.68 & 75.55 & 93.80 & \textbf{90.80} & \textbf{73.28} & 43.49 & 25.02 & \textbf{20.73} & 26.98 \\
\rowcolor{softcyan}
$\mathbf{32}$ & \textbf{91.60} & 75.34 & \textbf{94.60} & 90.69 & 73.27 & 44.46 & \textbf{25.09} & 20.66 & 26.88 \\
$64$ & 90.38 & \textbf{76.25} & 94.00 & 90.26 & 73.19 & \textbf{44.49} & 25.00 & 20.69 & \textbf{27.01} \\
\bottomrule
\end{tabular}%
}

\end{table*}

\paragraph{Results.}
The default $n_q=32$ achieves the highest Total Score ($81.68$), improving over $n_q=16$ and $n_q=64$ by $0.14$ and $0.23$ points. It also gives the highest Object Class ($91.60$), Human Action ($94.60$), and Dynamic Degree ($61.39$) scores. Temporal Flickering is identical at $99.02$, Motion Smoothness varies by only $0.01$ points, and chunk discontinuity ranges from $3.54$ to $3.55$. Other dimensions favor different query limits: $n_q=64$ improves Multiple Objects, Background Consistency, Scene, and Overall Consistency, whereas $n_q=16$ improves Aesthetic Quality, Color, Spatial Relation, and Appearance Style.

\paragraph{Efficiency and default.}
The measured throughputs are $9.93$, $10.00$, and $9.78$ FPS for $n_q=16$, $32$, and $64$, respectively. Their small differences do not isolate the cost of query selection and may include system-level variation. We use $n_q=32$ because it provides the highest measured aggregate quality without a throughput penalty in this comparison, not because it is optimal for every semantic dimension.

\captionsetup[figure]{font=small,skip=4pt}
\setlength{\floatsep}{10pt plus 2pt minus 2pt}
\setlength{\textfloatsep}{10pt plus 2pt minus 2pt}
\setlength{\intextsep}{10pt plus 2pt minus 2pt}
\setlength{\dblfloatsep}{10pt plus 2pt minus 2pt}
\setlength{\dbltextfloatsep}{10pt plus 2pt minus 2pt}
\makeatletter
\setlength{\@fptop}{0pt}
\setlength{\@fpsep}{12pt plus 2pt minus 2pt}
\setlength{\@fpbot}{0pt plus 1fil}
\setlength{\@dblfptop}{0pt}
\setlength{\@dblfpsep}{12pt plus 2pt minus 2pt}
\setlength{\@dblfpbot}{0pt plus 1fil}
\makeatother

\section{Additional Semantic and Temporal Analyses}
\label{app:additional_analyses}

These analyses reuse existing VBench videos, evaluator outputs, and fixed planner decompositions. They assess the distribution and temporal behavior of the reported semantic differences without introducing additional generation experiments. Uncertainty is estimated over prompts; seeds from the same prompt are kept together when multiple seeds are available.

\subsection{Paired Sample-Level Reliability}
\label{app:paired_reliability}

\begin{figure*}[htbp]
    \centering
    \includegraphics[width=0.96\linewidth,height=0.24\textheight,keepaspectratio]{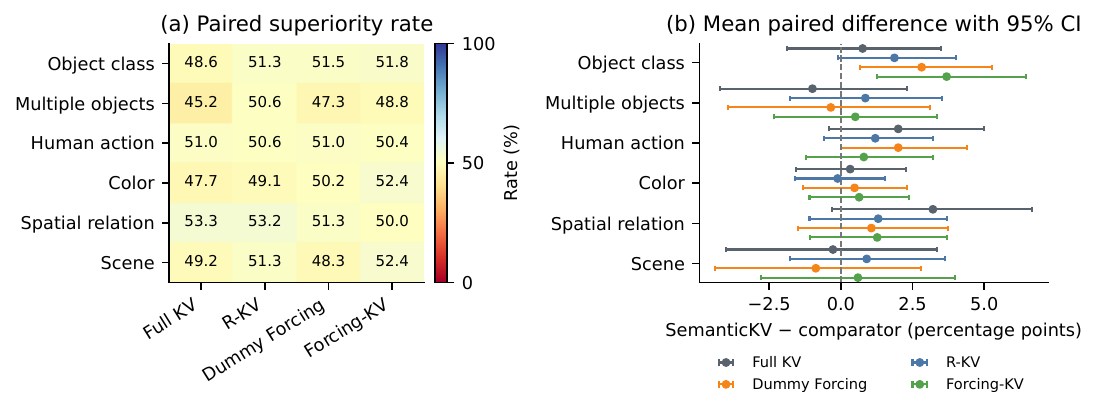}
    \caption{Paired semantic fidelity on matched prompt--seed tuples. (a) Tie-aware superiority rates. (b) Mean paired differences with prompt-clustered bootstrap 95\% confidence intervals. Positive differences favor \method{}; Full KV is the uncompressed reference.}
    \label{fig:paired_reliability}
\end{figure*}

We pair each available prompt--seed tuple between \method{} and a comparator for six semantic dimensions. Each dimension--comparator pair contains 395--500 videos from 79--100 unique prompts. We average the five seed-level differences within each prompt and resample prompts 10,000 times for cluster-bootstrap 95\% confidence intervals. Figure~\ref{fig:paired_reliability} reports both the mean differences and the tie-aware superiority rate $P(\Delta>0)+\frac{1}{2}P(\Delta=0)$, where $\Delta$ is the paired score difference. Table~\ref{tab:paired_reliability} lists the differences; its macro column averages the six dimensions without weighting and is not the official VBench Semantic Score. The reported intervals are marginal, not simultaneous multiple-comparison guarantees.

\paragraph{Paired semantic analysis.}
On matched prompt--seed outputs (Figure~\ref{fig:paired_reliability}), the macro-average differences across six semantic dimensions favor \method{} over R-KV, Dummy Forcing, and Forcing-KV by $1.00$, $0.86$, and $1.25$ points. Individual dimensions remain mixed, and most confidence intervals include zero. The following subsections examine semantic complexity and temporal drift; Appendix~\ref{app:vbench_qualitative} provides qualitative examples.

\begin{table*}[htbp]
\centering
\captionsetup{skip=3pt}
\small
\setlength{\tabcolsep}{7pt}
\caption{Mean paired semantic-score difference (\method{} minus comparator), in percentage points. The macro column averages the six dimensions without reweighting them.}
\label{tab:paired_reliability}
\resizebox{\textwidth}{!}{%
\begin{tabular}{@{}lrrrrrrr@{}}
\toprule
\textbf{Comparator} & \textbf{Object} & \textbf{Multiple objects} & \textbf{Human action} & \textbf{Color} & \textbf{Spatial relation} & \textbf{Scene} & \textbf{Macro} \\
\midrule
Full KV       & $+0.76$ & $-0.99$ & $+2.00$ & $+0.32$ & $+3.21$ & $-0.28$ & $+0.84$ \\
R-KV          & $+1.87$ & $+0.85$ & $+1.20$ & $-0.11$ & $+1.30$ & $+0.90$ & $+1.00$ \\
Dummy Forcing & $+2.82$ & $-0.35$ & $+2.00$ & $+0.48$ & $+1.06$ & $-0.87$ & $+0.86$ \\
Forcing-KV    & $+3.69$ & $+0.50$ & $+0.80$ & $+0.64$ & $+1.27$ & $+0.60$ & $+1.25$ \\
\bottomrule
\end{tabular}}
\end{table*}

The macro differences are positive against all three compressed baselines. Of the 18 dimension--baseline comparisons, 15 favor \method{}, including all six comparisons with Forcing-KV. Object Class improves by $2.82$ points over Dummy Forcing and $3.69$ over Forcing-KV, with intervals excluding zero. The exceptions are Multiple Objects and Scene against Dummy Forcing ($-0.35$ and $-0.87$ points), and Color against R-KV ($-0.11$). Most remaining dimension-level intervals include zero. The evidence therefore supports a positive macro trend without establishing significance for every dimension.

\subsection{Performance versus Semantic Complexity}
\label{app:semantic_complexity}

\begin{figure}[!htbp]
    \centering
    \includegraphics[width=0.6\linewidth,height=0.28\textheight,keepaspectratio]{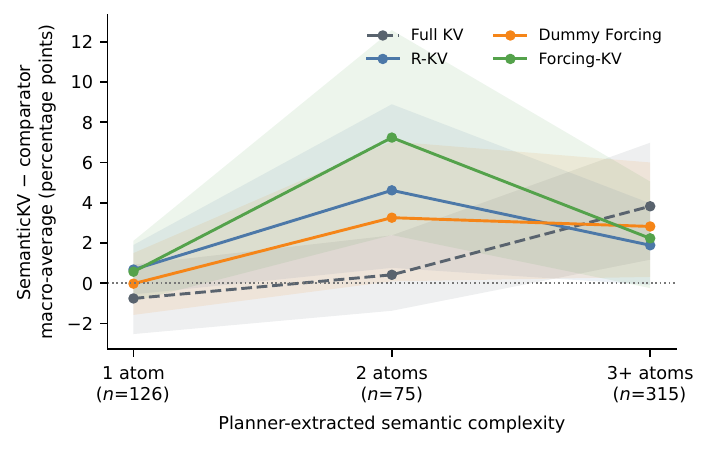}
    \caption{\textbf{Semantic fidelity as a function of planner-extracted prompt complexity.} Values are macro-averaged paired differences between \method{} and each comparator across six semantic VBench dimensions. Shaded regions show prompt-clustered bootstrap 95\% confidence intervals. The grouping uses the planner decomposition only and is fixed before inspecting the video scores.}
    \label{fig:semantic_complexity}
\end{figure}

The 946 benchmark entries contain 944 distinct prompt strings, for which the planner outputs fixed decompositions. Among prompts evaluated on the six semantic dimensions, we group them by one, two, or at least three semantic constraints, yielding 126, 75, and 315 unique prompts. Grouping depends only on the planner output, not on evaluation scores. Within each group, we compute paired prompt-level differences per dimension and macro-average the six dimensions; uncertainty uses the same 10,000-sample prompt-clustered bootstrap protocol as above. The ``atoms'' shown in Figure~\ref{fig:semantic_complexity} refer to these semantic constraints.

For one-constraint prompts, the macro differences against R-KV, Dummy Forcing, and Forcing-KV are $+0.68$, $-0.02$, and $+0.58$ points. They rise to $+4.62$, $+3.26$, and $+7.24$ for two constraints, and remain positive at $+1.89$, $+2.82$, and $+2.23$ for at least three constraints. All two-constraint intervals exclude zero. In the largest group, the interval excludes zero against Dummy Forcing but narrowly crosses zero against R-KV and Forcing-KV. Relative to Full KV, the corresponding differences are $-0.75$, $+0.42$, and $+3.83$, with the last interval $[+1.17,+6.98]$. The pattern suggests a benefit for compositional prompts, but is not monotonic in constraint count.

\subsection{Within-Video Temporal Semantic Drift}
\label{app:temporal_drift}

This diagnostic reuses existing videos rather than generating new samples. We select the first 24 lexicographically sorted prompts from each of Object Class, Multiple Objects, Human Action, and Scene for which all five methods are available, using seed 42. The fixed 96-prompt subset contains 480 videos. Five timestamps---0\%, 25\%, 50\%, 75\%, and 100\% of normalized duration---give 2,400 evaluated frames. CLIP ViT-B/32~\citep{radford2021clip} measures image--prompt alignment and visual-feature similarity to the first frame. Alignment is reported as a change from the first frame to reduce prompt-specific offsets; confidence intervals are obtained by resampling prompts.

\begin{figure*}[!htbp]
    \centering
    \includegraphics[width=0.92\linewidth,height=0.24\textheight,keepaspectratio]{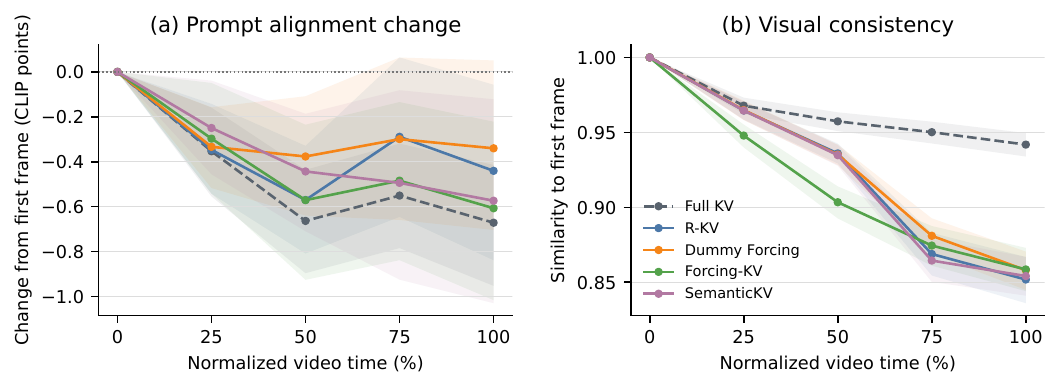}
    \caption{\textbf{Within-video drift on a fixed 96-prompt subset.} (a) Change in CLIP image--prompt alignment relative to the first frame. (b) CLIP visual-feature similarity to the first frame. Curves show means and shaded prompt-bootstrap 95\% confidence intervals over 96 videos per method. Full KV is the uncompressed reference.}
    \label{fig:temporal_semantic_drift}
\end{figure*}

\begin{table}[!htbp]
\centering
\captionsetup{skip=3pt}
\small
\setlength{\tabcolsep}{6pt}
\caption{End-of-video temporal analysis. Brackets give prompt-bootstrap 95\% confidence intervals. A less negative alignment change and a higher visual similarity indicate less drift.}
\label{tab:temporal_drift}
\begin{tabular}{@{}lrr@{}}
\toprule
\textbf{Method} & \shortstack{\textbf{Alignment change}\\\textbf{(CLIP points)}} & \shortstack{\textbf{Visual similarity}\\\textbf{to first frame}} \\
\midrule
Full KV       & $-0.67\;[-0.95,-0.41]$ & $0.942\;[0.934,0.950]$ \\
R-KV          & $-0.44\;[-0.84,-0.06]$ & $0.852\;[0.836,0.867]$ \\
Dummy Forcing & $-0.34\;[-0.70,+0.05]$ & $0.858\;[0.845,0.870]$ \\
Forcing-KV    & $-0.61\;[-1.02,-0.22]$ & $0.859\;[0.844,0.873]$ \\
\rowcolor{softcyan}
\textbf{\method{}} & $-0.57\;[-1.03,-0.12]$ & $0.854\;[0.841,0.867]$ \\
\bottomrule
\end{tabular}
\end{table}

All methods show a modest decline in prompt alignment. At the final frame, the paired alignment differences between \method{} and R-KV, Dummy Forcing, and Forcing-KV are $-0.13$, $-0.23$, and $+0.03$ CLIP points; all paired intervals include zero. Paired visual-similarity differences are at most $0.0044$, again with intervals including zero. Full KV has higher visual similarity than every compressed method: \method{} is $0.088$ lower, with an interval excluding zero. Because valid object or camera motion can also reduce feature similarity, this is a drift diagnostic rather than a standalone quality ranking. We find no detectable difference between the compressed methods on this subset, which should not be interpreted as proof of temporal equivalence.

\section{Qualitative Comparisons}
\label{app:vbench_qualitative}

\begin{figure*}[!htbp]
    \centering
    \includegraphics[width=0.90\linewidth,height=0.37\textheight,keepaspectratio]{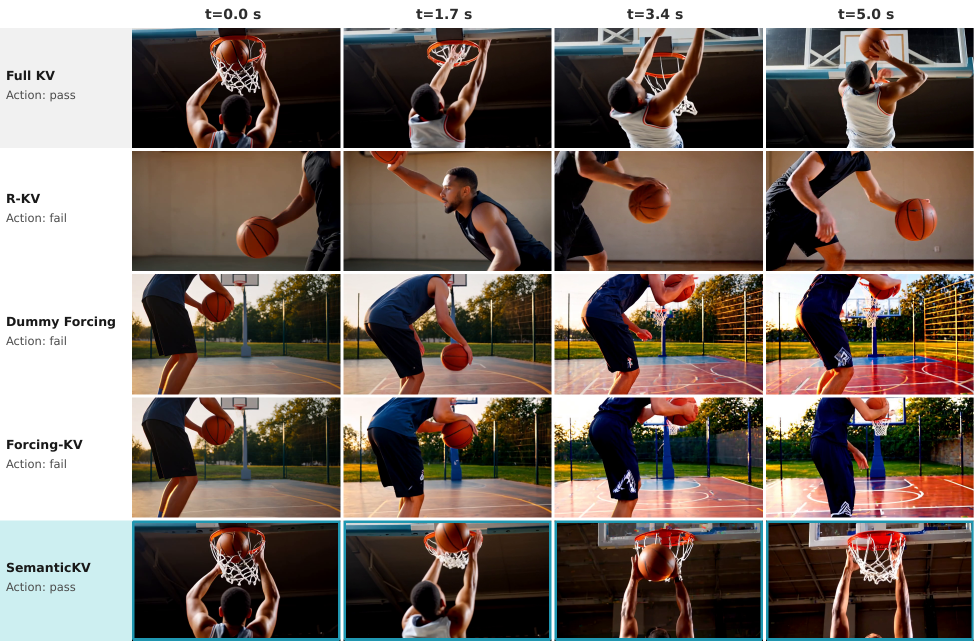}
    \caption{\textbf{Human-action fidelity on LongLive.} Prompt: ``A person is dunking basketball'' (seed 42). R-KV, Dummy Forcing, and Forcing-KV depict ball handling or dribbling and fail the VBench Human Action check. \method{} depicts the requested dunk and passes the check, matching the Full-KV reference. Rows compare methods and columns show uniformly sampled timestamps.}
    \label{fig:qualitative_action}
\end{figure*}

\begin{figure*}[!htbp]
    \centering
    \includegraphics[width=0.90\linewidth,height=0.37\textheight,keepaspectratio]{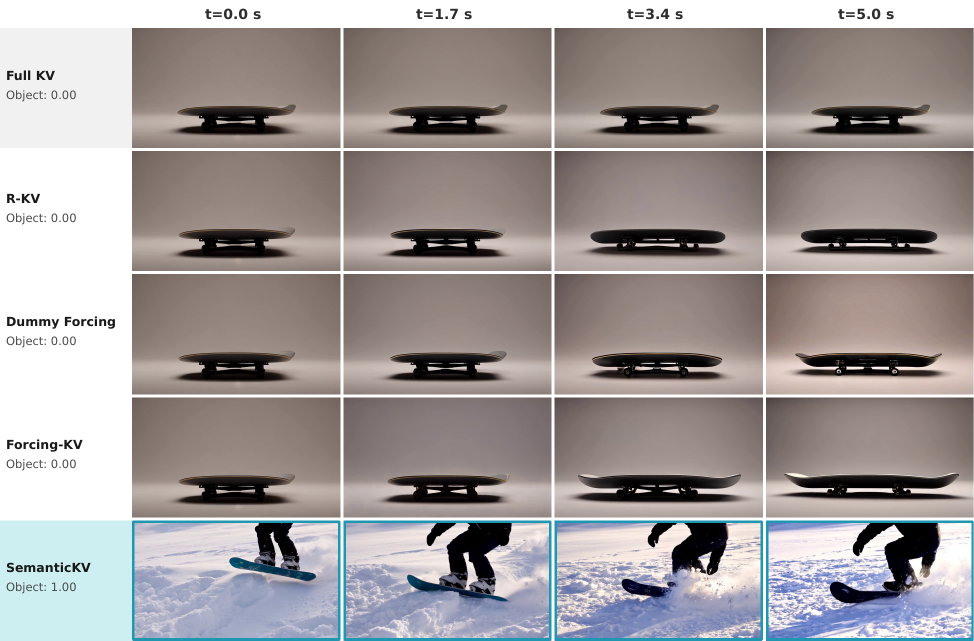}
    \caption{\textbf{Object-class fidelity on LongLive.} Prompt: ``a snowboard'' (seed 45). Full KV and all three compressed baselines generate a skateboard against a studio background and receive a per-video Object Class score of 0.00. \method{} generates a snowboard being ridden on snow and receives 1.00. The example illustrates that semantically guided retention can preserve the requested class even when a visually related concept is a strong failure mode.}
    \label{fig:qualitative_single_object}
\end{figure*}

\begin{figure*}[!htbp]
    \centering
    \includegraphics[width=0.90\linewidth,height=0.37\textheight,keepaspectratio]{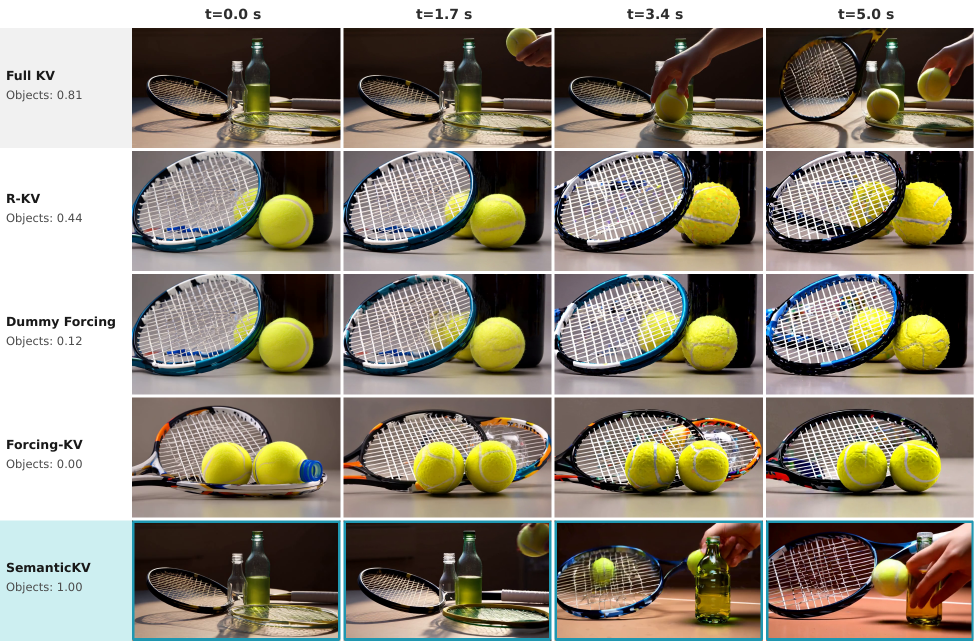}
    \caption{\textbf{Multi-object fidelity on LongLive.} Prompt: ``a tennis racket and a bottle'' (seed 43). The compressed baselines replace the requested bottle with tennis balls, producing per-video Multiple Objects scores of 0.44, 0.12, and 0.00 for R-KV, Dummy Forcing, and Forcing-KV, respectively. \method{} retains both requested objects and scores 1.00; Full KV scores 0.81.}
    \label{fig:qualitative_objects}
\end{figure*}

\begin{figure*}[!htbp]
    \centering
    \includegraphics[width=0.90\linewidth,height=0.37\textheight,keepaspectratio]{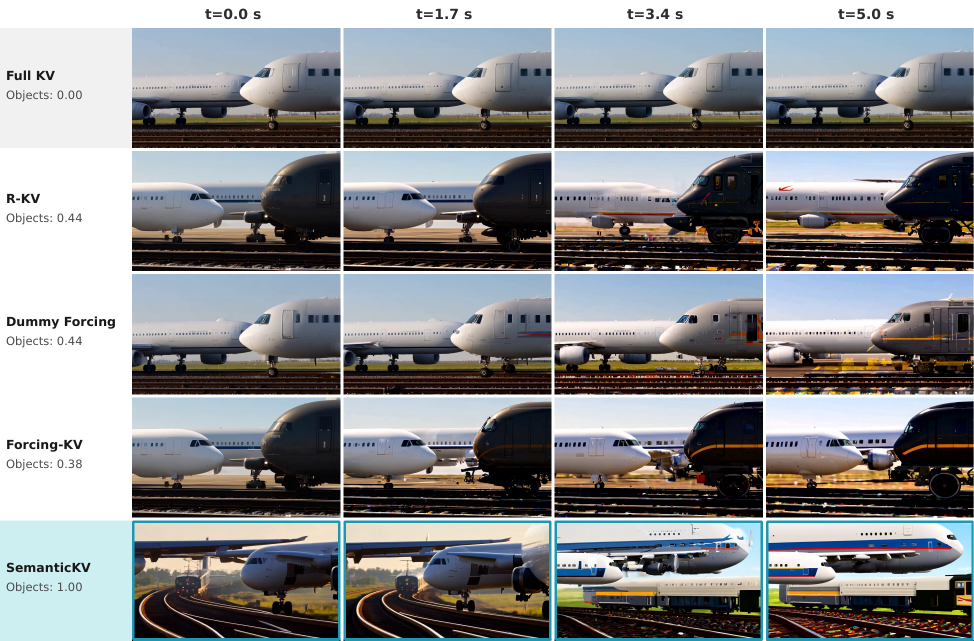}
    \caption{\textbf{Compositional object fidelity on LongLive.} Prompt: ``an airplane and a train'' (seed 44). The compressed baselines produce ambiguous airplane--train hybrids and obtain per-video Multiple Objects scores of 0.44, 0.44, and 0.38 for R-KV, Dummy Forcing, and Forcing-KV, respectively. \method{} keeps an airplane and a rail vehicle visually distinguishable and scores 1.00; Full KV scores 0.00.}
    \label{fig:qualitative_airplane_train}
\end{figure*}

\begin{figure*}[!htbp]
    \centering
    \includegraphics[width=0.90\linewidth,height=0.37\textheight,keepaspectratio]{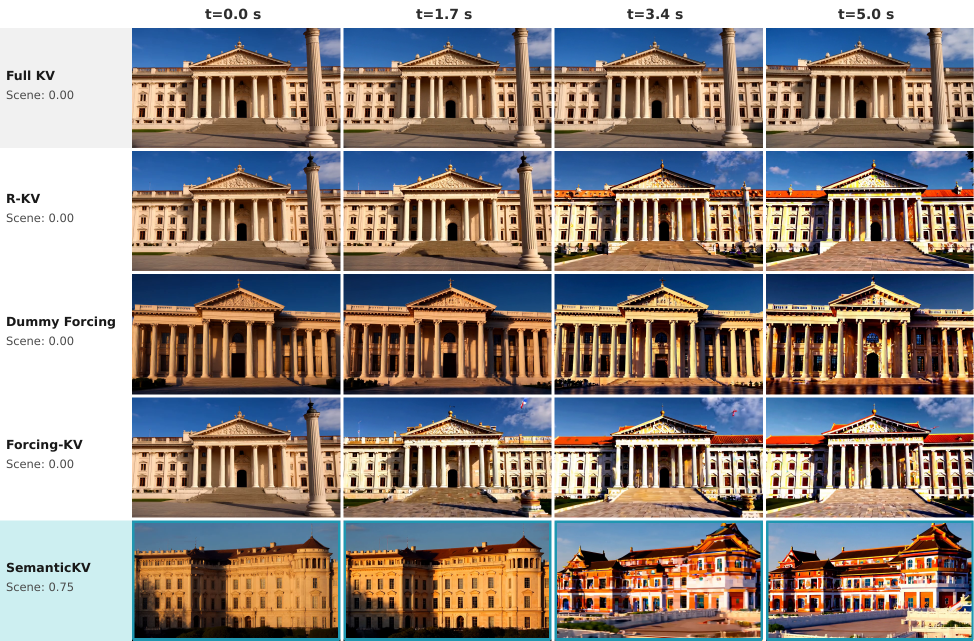}
    \caption{\textbf{Scene fidelity on LongLive.} Prompt: ``palace'' (seed 45). Full KV and the three compressed baselines converge to a generic neoclassical institutional facade and receive per-video Scene scores of 0.00. \method{} produces palatial architecture across the sequence and scores 0.75.}
    \label{fig:qualitative_palace}
\end{figure*}

\begin{figure*}[!htbp]
    \centering
    \includegraphics[width=0.90\linewidth,height=0.37\textheight,keepaspectratio]{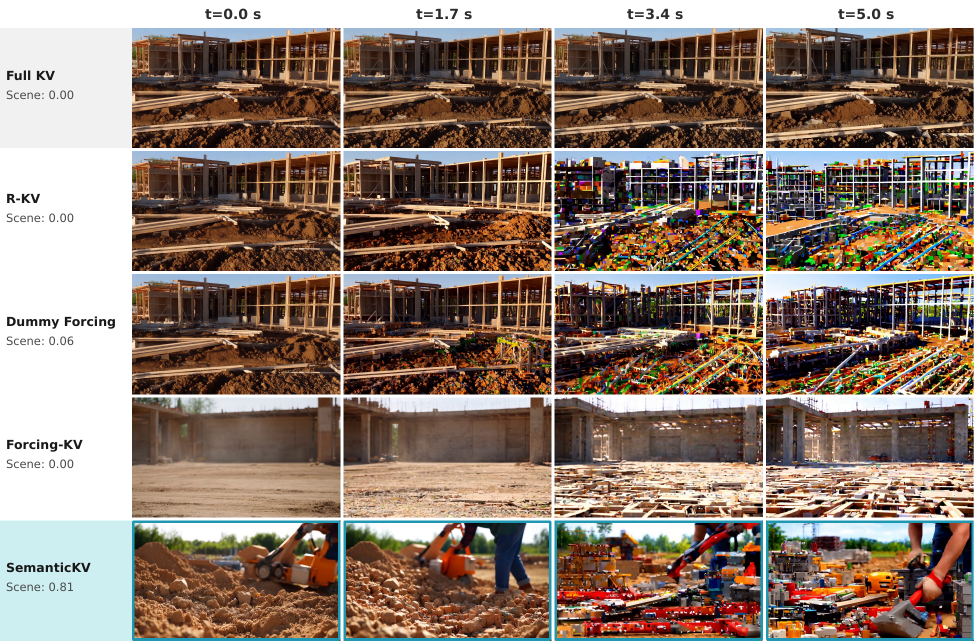}
    \caption{\textbf{Activity-rich scene fidelity on LongLive.} Prompt: ``construction site'' (seed 43). \method{} preserves discriminative construction cues, including excavation, a worker, tools, and machinery, and obtains a per-video Scene score of 0.81. The corresponding scores are 0.00 for Full KV, 0.00 for R-KV, 0.06 for Dummy Forcing, and 0.00 for Forcing-KV.}
    \label{fig:qualitative_construction_site}
\end{figure*}

Figures~\ref{fig:qualitative_action}--\ref{fig:qualitative_construction_site} show diagnostic LongLive examples covering actions, object classes, multiple objects, and scenes. Methods are matched by original prompt and seed, and columns use the same uniformly spaced timestamps. The score beside each row is the official per-video result for the semantic dimension targeted by that prompt. Examples are selected where \method{} scores higher than every compressed baseline; they illustrate failure modes rather than estimate their frequency. The numerical comparisons in Table~\ref{tab:main_result_short_full} and the paired analyses in Appendix~\ref{app:paired_reliability}, including unfavorable dimensions, provide the quantitative context for these examples.

\end{document}